\documentclass[letterpaper]{article}
\usepackage{uai2019}
\usepackage[margin=1in]{geometry}

\usepackage{microtype}
\usepackage{graphicx}
\usepackage{booktabs} 

\usepackage{natbib}
\usepackage{amssymb, amsmath,amsthm}
\usepackage{amsfonts}
\usepackage{algorithm}
\usepackage{algorithmic}
\usepackage{paralist}
\usepackage{multirow}
\usepackage{times}
\usepackage{wrapfig}

\usepackage{url,enumerate}
\usepackage{color,xcolor}
\usepackage{makeidx}  
\usepackage{amsmath,amssymb}
\usepackage[small, compact]{titlesec}
\usepackage{xspace}
\usepackage{epstopdf}
\usepackage{mathrsfs}
\usepackage{times}
\usepackage{enumerate}
\usepackage{color}
\usepackage{graphicx,epsfig}
\usepackage{amsmath,amssymb,xspace}
\usepackage{url}
\usepackage{hyperref}

\usepackage{bm}
\usepackage{dsfont}
\usepackage{upgreek}
\usepackage{cleveref}
\usepackage{multirow}
\usepackage{ulem}
\usepackage{subcaption}
\newtheorem{theorem}{Theorem}[section]

\newtheorem{lem}[theorem]{Lemma}
\usepackage{hyperref}

\newcommand{\tr}{{\rm tr}}

\newcommand{\m}{{\bf m}}

\renewcommand{\u}{{\bf u}}
\newcommand{\w}{{\bf w}}
\newcommand{\x}{{\bf x}}
\newcommand{\y}{{\bf y}}
\newcommand{\z}{{\bf z}}
\newcommand{\A}{{\bf A}}

\newcommand{\I}{{\bf I}}

\newcommand{\N}{\mathcal{N}}  

\renewcommand{\S}{{\bf S}}

\newcommand{\U}{{\bf U}}

\newcommand{\X}{{\bf X}}

\newcommand{\Ycal}{{\mathcal{Y}}}

\newcommand{\Ucal}{{\mathcal{U}}}
\newcommand{\Dcal}{{\mathcal{D}}}
\newcommand{\Ocal}{{\mathcal{O}}}

\newcommand{\boldeta}{\boldsymbol{\eta}}

\newcommand{\blambda}{\boldsymbol{\lambda}}

\newcommand{\btheta}{\boldsymbol{\theta}}

\newcommand{\bSigma}{\boldsymbol{\Sigma}}

\newcommand{\1}{{\bf 1}}
\newcommand{\0}{{\bf 0}}

\newcommand{\ben}{\begin{enumerate}}
\newcommand{\een}{\end{enumerate}}

\newcommand{\ie}{{{i.e.,}}\xspace}
\newcommand{\eg}{{\textit{e.g.,}}\xspace}
\newcommand{\EE}{\mathbb{E}}

\newcommand{\cmt}[1]{}
\newcommand{\tf}{\tilde{f}}

\newcommand{\bi}{{\bf i}}
\newcommand{\bj}{{\bf j}}

\title{Conditional Expectation Propagation}

\cmt{
\author{ {\bf Zheng Wang} \\
School of Computing \\
University of Utah\\
Salt Lake City, UT 84112 \\
\And
{\bf Shandian Zhe}  \\
School of Computing \\
University of Utah\\
Salt Lake City, UT 84112 \\
}
}
\author{
	Zheng Wang, Shandian Zhe\\
	School of Computing, University of Utah\\
	Salt Lake City, UT 84112\\	
	\texttt{wzhut@cs.utah.edu}, \texttt{zhe@cs.utah.edu}\\
}

\begin{document}

\maketitle

\begin{abstract}  
Expectation propagation (EP) is a powerful approximate inference algorithm. However, a critical barrier in applying EP is that the moment matching in message updates can be intractable. 
Handcrafting approximations is usually tricky, and lacks generalizability. Importance sampling is very expensive. While Laplace propagation provides a good solution, it has to run numerical optimizations to find Laplace approximations in every update, which is still quite inefficient. To overcome these practical barriers, we propose conditional expectation propagation (CEP) that performs conditional moment matching given the variables outside each message, and then takes expectation w.r.t the approximate posterior of these variables. The conditional moments are often analytical and much easier to derive. In the most general case, we can use (fully) factorized messages to represent the conditional moments by quadrature formulas. We then compute the expectation of the conditional moments via Taylor approximations when necessary.  In this way, our algorithm can always conduct efficient, analytical fixed  point iterations. Experiments on several popular models for which standard EP is available or unavailable demonstrate the advantages of CEP in both inference quality and computational efficiency.

\end{abstract}



\section{INTRODUCTION}

Expectation propagation (EP)~\citep{minka2001expectation} is a popular posterior inference algorithm. It approximates the factors of the joint probability with exponential-family terms, also called messages (in graphical models), and iteratively updates each message via moment matching. EP often produces fast and accurate posterior estimations,\cmt{, and is much more efficient than sampling based approaches. EP} and have been applied in many Bayesian learning tasks~\citep{NIPS2006_3079, graepel2010web,hernandez2015probabilistic}. EP is the cornerstone of the influential machine learning library, Infer.NET~\citep{minka2014infer}. 


Nonetheless, a critical barrier of using EP is that the moment matching can be intractable during the message updates. When a factor (\eg the likelihood of a data point) is complex and includes many latent variables, the normalizer of the tilted distribution (that is proportional to the factor multiplying with the messages from the other factors) is likely to be intractable and so are the moments. Handcrafting an approximation to the moments is usually tricky, and limited to particular types of factors. One can consider importance sampling, which, however, requires a large number of samples to obtain reliable estimations and hence is very expensive. 

To mitigate this issue, \citet{eskin2004laplace} proposed Laplace propagation (LP) that uses Gaussian messages and calculates the Laplace approximation to the tilted distribution. This is fulfilled by finding the mode and computing the inverse Hessian\cmt{of the logarithm of the unnormalized tilted distribution} at the mode. The approximation, which is a Gaussian distribution, is then used for moment matching.   Despite its effectiveness, LP has to repeatedly conduct numerical optimizations to find the modes in message updates, and therefore is still quite inefficient. 


To overcome these practical barriers, we propose conditional expectation propagation (CEP) algorithm, which not only bypasses the (possibly) tricky moment matching and so the nontrivial approximation designs, but also enjoys efficient and analytical fixed point iterations. Specifically, we observe that although the moments of the tilted distributions can be intractable, the conditional moments given a set of variables fixed  are often tractable and analytical. In the most general case, we can consider the conditional moments for a single variable (or two) given all the others. These  moments (if unavailable) can be explicitly represented by  quadrature formulas. Therefore, we first introduce factorized approximations (\ie messages)  and match the conditional moments for the variable(s) in each message. Next, we update the message by computing the expectation of the conditional moments w.r.t the (current) approximate posterior of the remaining variables. \cmt{We show how this idea links to the standard EP.}We show the connection of this method to the standard EP. Finally, if necessary, to enable tractable expectation computation for the conditional moments, we use their first or second order Taylor approximations at the moments of the remaining variables. In this way, CEP always performs efficient, analytical updates and is general to all kinds of factors, without nontrivial approximation designs, numerical optimizations or importance sampling.

For evaluation, we first examined CEP in Bayesian probit and logistic regression where the standard moment matching is analytical or has accurate approximations. For logistic regression, we represented the conditional moments with Gauss-Hermite quadrature. On both simulation and real-world datasets, CEP obtains  inference quality and running efficiency close to EP, and is much faster than LP and importance sampling based approaches.  We then applied CEP in Bayesian tensor decomposition where the moment matching is intractable. CEP largely improves upon LP and a classical tensor decomposition algorithm, and is close to or significantly better than the variational message passing (VMP) in prediction accuracy. Meanwhile,  CEP achieves over $40$x speedup against LP. Finally,  we adapted CEP to the assumed density filtering (ADF) framework to perform streaming Bayesian decomposition. Our method  outperforms the state-of-the-art approach~\citep{du2018probabilistic} by a large margin. 


\section{EXPECTATION PROPAGATION}
We first review the expectation propagation (EP) algorithm~\cite{minka2001expectation}. 
Given the observations $\Dcal$, the posterior distribution of a Bayesian model has a general form,
\begin{align}
p(\btheta|\Dcal) = \frac{1}{Z} \prod\nolimits_{i} f_i(\btheta_i) \label{eq:post}
\end{align}
where $\btheta$ are the latent random variables, $Z$ the normalization constant, and $\{f_i(\btheta_i)\}_i$ the factors that link to the prior or data likelihoods. For example, $f_0$ may correspond to the prior, and $f_n$ the likelihood of the $n$-th data point ($n\ge 1$).  Each $\btheta_i$ is a subset of or identical to $\btheta$.  Bayesian inference aims to compute the posterior distribution $p(\btheta|\Dcal)$. However, the exact computation is usually infeasible, due to the high dimensional and intractable integral in calculating the normalizer $Z$. 

To address this issue, EP  approximates each factor $f_i$ by an exponential-family term,
 \begin{align}
 \tf_i(\btheta_i)\propto \exp\big(\blambda_i^\top \phi(\btheta_i)\big),
 \end{align}
 where $\blambda_i$ and $\phi(\btheta_i)$ are the natural parameters and sufficient statistics, respectively. Then the approximate posterior distribution is given by 
\begin{align}
q(\btheta) \propto \prod_i \tf_i(\btheta_i), 
\end{align}
and trivial to compute through the summation of the natural parameters. In the factor graph representation~\citep{kschischang2001factor},  the approximation factor $\tf_i(\btheta_i)$ is also defined as the message from  factor $f_i$ to the variables $\btheta_i$. Commonly used messages are Gaussian factors. 

\cmt{Starting with an initialization, }EP repeatedly refines each approximation factor $\tf_i$  through three steps, message deletion, projection and update.  In the message deletion step, we compute a calibrating distribution $q^{\setminus i}(\btheta)$ by removing $\tf_i$ from the current posterior $q(\btheta)$, $q^{\setminus i}(\btheta) \propto q(\btheta)/\tf_i(\btheta_i)$.  This is equivalent to multiplying the messages from all the other factors, and hence a surrogate for these context factors.  In the projection step, we first construct a tilted distribution $\hat{p}_i(\btheta)$ from multiplying the true factor $f_i$ with the calibrating distribution, 
\begin{align}
\hat{p}_i(\btheta) = \frac{1}{Z_i} f_i(\btheta_i)q^{\setminus i}(\btheta), \label{eq:tilt}
\end{align}
where $Z_i$ is the normalizer. We then project the tilted distribution to the exponential family by minimizing the KL divergence between $\hat{p}_i(\btheta)$ and a new approximate posterior $q^*(\btheta)$.
It is well known that the minimum can be obtained by moment matching. That is, we compute the expectation of $\phi(\btheta)$ w.r.t $\hat{p}_i$ and set it to the moment of $q^*(\btheta)$, 
\begin{align}
\EE_{q^*}\big(\phi(\btheta)\big) = \EE_{\hat{p}_i} \big(\phi(\btheta)\big) = \nabla_{\blambda^{\setminus i}}\log(Z_i) \label{eq:mm}
\end{align}
 where $\blambda^{\setminus i}$ are the natural parameters of  the calibrating distribution $q^{\setminus i}$. 
Finally, we update the message $\tf_i$ via $\tf_i(\btheta_i) \propto q^*(\btheta)/q^{\setminus i}(\btheta)$.

The refinement of each message can be sequential or parallel. The key step in the refinement is the moment matching \eqref{eq:mm}. Through repeated moment matching, the approximate posterior is improved by assimilating the critical statistics in the true posterior. Due to the fixed point iteration nature, EP often converges fast (although convergence is not always guaranteed) and presents prominent computational advantages over alternative approaches, such as sampling.  


\section{CONDITIONAL EXPECTATION PROPAGATION}
Despite the impressive success of EP,  applying EP can be troublesome when the moment matching is intractable. Given a complex factor $f_i(\btheta_i)$, it is very likely that the log normalizer of the tilted distribution, $\log(Z_i)$, is intractable and hence the moment calculation (see \eqref{eq:mm}). In this case, we might need to handcraft an approximation of the moments. However, this is often tricky and the approximation is hard to generalize to other types of complex factors. A simple and general approach is to use importance sampling, but it requires a large number of samples to obtain a reliable estimation, hence is very costly and will deprive the computational advantage of EP. Although this problem can be alleviated by training a machine learning model to predict the moments~\citep{heess2013learning,jitkrittum2015kernel}, the training data collection and replenishment still require a lot of importance sampling. In addition, learning predictors for distinct kinds of factors or even the same kind with new observations may also require us to supplement training samples and/or retrain from scratch~\citep{jitkrittum2015kernel}. 
While Laplace propagation~\citep{eskin2004laplace} can completely sidestep importance sampling and provide a general solution by using Gaussian messages and calculating Laplace approximation to the titled distribution, it has to run numerical optimizations, say, L-BFGS, to find the mode and to construct the approximation in every message update, which is still quite inefficient. 


To overcome these  barriers, avoiding nontrivial approximation designs, costly importance sampling and numerical optimizations,  we propose the conditional expectation propagation (CEP) algorithm,  presented as follows.   

\subsection{CONDITIONAL MOMENT MATCHING}
We observe that while the moments of the tilted distribution~\eqref{eq:tilt} can be intractable, the conditional moments of a subset of variables given the others can be analytical and easy to derive. For example,  when a tilted distribution is proportional to a production factor multiplying with a Gaussian distribution, \ie   $\hat{p}_i(\btheta) \propto \N(x_n | \btheta_{1}^\top \btheta_{2}, \beta) \N(\btheta|\mu, \bSigma)$ where $\btheta = [\btheta_1^\top, \btheta_2^\top]^\top$, computing the moments of $\btheta$ is tricky, but $\btheta_1$ given $\btheta_2$ fixed and $\btheta_2$ given $\btheta_1$ fixed is trivial --- both $\hat{p}(\btheta_1|\btheta_2)$ and $\hat{p}(\btheta_2|\btheta_1)$ are Gaussian distributions. This resembles Gibbs sampling --- while the posterior is intractable, the conditional distribution of each variable given all the others (and observations) is tractable and we can repeatedly sample from the conditional distributions. 

Therefore, we introduce factorized messages and derive analytical conditional moments in the first step. W.l.o.g, we partition $\btheta$ into $\{\btheta_1, \ldots, \btheta_M\}$ and approximate each factor $f_i$ with $\prod_m \tf_{im}(\btheta_m)$ where each $\tf_{im}(\btheta_m) \propto \exp\big(\blambda_{im}^\top \phi(\btheta_{m})\big)$. Hence, the approximate posterior $q(\btheta)$ and the calibrating distribution $q^{\setminus i}(\btheta)$ are both factorized over $\{\btheta_1, \ldots, \btheta_M\}$.  Note that factorized messages are widely adopted by EP and other message passing algorithms, especially in large-scale applications~\citep{graepel2010web,zhe2016online}.  To update each $\tf_{im}$ in the standard EP, we need to compute the moments of $\btheta_m$ with the tilted distribution $\hat{p}_i(\btheta) \propto q^{\setminus i}(\btheta_m) q^{\setminus i}(\btheta_{\setminus m}) f_i(\btheta_{m}, \btheta_{\setminus m})$
where $\btheta_{\setminus m}$ are all the latent variables excluding $\btheta_m$. It can be seen that
\begin{align}
\EE_{\hat{p}_i(\btheta)}\big(\phi(\btheta_{m})\big) = \EE_{\hat{p}_i(\btheta_{\setminus m})}\big[ \EE_{\hat{p}_i(\btheta_{m}|\btheta_{\setminus m})} \big(\phi(\btheta_{m})\big)\big] \label{eq:expt}
\end{align}
where $\hat{p}_i(\btheta_{\setminus m})$ is $\hat{p}_i(\btheta)$ marginalizing out $\btheta_m$, and 
\begin{align}
\hat{p}_i(\btheta_{m}|\btheta_{\setminus m}) \propto q^{\setminus i}(\btheta_{m}) f_i(\btheta_{m}, \btheta_{\setminus m}). \label{eq:cond-tilt}
\end{align}
We first obtain an analytical form of the conditional moment $\EE_{\hat{p}_i(\btheta_{m}|\btheta_{\setminus m})} \big(\phi(\btheta_{m})\big)$. This is straightforward for many factors. But uniformly,  this can always be achieved by using fully factorized messages, and representing the conditional moment (which is now for a single variable) with a quadrature formula, $\EE_{\hat{p}_i(\theta_{m}|\btheta_{\setminus m})} \big(\phi(\theta_{m})\big) \approx \sum_j \alpha_j g(\gamma_j, \btheta_{\setminus m})$ where $\{\gamma_j, \alpha_j\}$ are quadrature nodes and weights. In addition, we can also use bi-variable or triple-variable messages with two-dimensional or three-dimensional quadrature formulas that are more complex. However, higher-dimensional quadratures are not recommended due to the degradation of accuracy and explosion of the computational cost.  

\subsection{EXPECTED CONDITIONAL MOMENT}
To update the message $\tf_{im}$, EP requires us to further compute the expectation of the conditional moment w.r.t the marginal tilted distribution $\hat{p}_i(\btheta_{\setminus m})$ (see \eqref{eq:expt}). However, $\hat{p}_i(\btheta_{\setminus m})$ can be intractable for tricky $f_i$ as well. To overcome this problem, we observe that (due to the factorized posterior form) EP also maintains the moment matching between $\hat{p}_i(\btheta_{\setminus m})$ and 
$q(\btheta_{\setminus m})$ --- the (marginal) approximate posterior for $\btheta_{\setminus m}$. Therefore, we can assume they are close in high density regions and use $q(\btheta_{\setminus m})$ as a surrogate for $\hat{p}_i(\btheta_{\setminus m})$. 
The conditional moment is a function of $\btheta_{\setminus m}$, and w.l.o.g can be further represented as a function of their sufficient statistics, $\EE_{\hat{p}_i(\btheta_{m}|\btheta_{\setminus m})}(\phi(\btheta_{m})) = h(\Phi_m)$ where $\Phi_m = \{\phi(\btheta_1), \ldots, \phi(\btheta_{m-1}), \phi(\btheta_{m+1}), \ldots, \phi(\btheta_{M})\} $. Therefore, we next compute the expected conditional moment $\EE_{q(\btheta_{\setminus m})} \big(h(\Phi_m)\big)$. 

Even with $q(\btheta_{\setminus m})$, the expectation might still be intractable. However, with the nice and easy form of $q(\btheta_{\setminus m})$,  we can use the Taylor expansion at the moments of $\btheta_{\setminus m}$, \ie $\EE_q (\Phi_m)$, to derive the first-order or second-order approximation to $h(\Phi_m)$, 
\begin{align}
&\hat{h}_1( \Phi_m ) =h( \EE_q (\Phi_m) ) + \nabla h( \EE_q (\Phi_m) ) ^\top \u  \label{eq:taylor1st}\\
&\hat{h}_2(\Phi_m) = \hat{h}_1(\Phi_m) + \frac{1}{2} \tr\big(\S \cdot \nabla \nabla h( \EE_q (\Phi_m) )\big) \label{eq:taylor2nd}
\end{align}
where $\u = \Phi_m - \EE_q (\Phi_m) $ and $\S = \u\u^\top$. Taking expectation over the Taylor approximations, we have 
\begin{align}
&\EE_{q(\btheta_{\setminus m})} \big(h(\Phi_m)\big) \approx h( \EE_q (\Phi_m) ) \;\;\; \mathrm{or}\; \label{eq:1st} \\
& h( \EE_q (\Phi_m) ) + \frac{1}{2} \tr\big(\mathrm{cov}(\Phi_m) \cdot \nabla \nabla h( \EE_q (\Phi_m) )\big), \label{eq:2nd}
\end{align}
Both \eqref{eq:1st} and \eqref{eq:2nd} are analytical and straightforward to compute. Using \eqref{eq:2nd} can be more accurate but requires extra calculations. To reduce the cost, we can use the diagonal Hessian and covariance matrix.

The other steps are the same as standard EP. As we can see, through deriving analytical conditional moments and their expectations, our method bypasses the tricky moment matching, and still conducts efficient and analytical message refinements. We do not need to run expensive importance sampling, numerical optimizations or design nontrivial moment approximations. The proposed CEP is summarized in Algorithm \ref{alg:cep}.  

\subsection{CONNECTION TO EP}
A key difference between CEP and EP is that CEP uses the approximate posterior $q(\btheta_{\setminus m})$ to replace the marginal tilted distribution $\hat{p}_i(\btheta_{\setminus m})$ in computing the expected conditional moments as in  \eqref{eq:expt}. This can lead to a different fixed point. But under certain conditions, EP's fixed point can still coincide with CEP's.  
\begin{lem}
	When the conditional moment $h$ is part of the sufficient statistics of $\btheta_{\setminus m}$, \ie each element of $h$ belongs to $\Phi_m$, the fixed points of EP are also that of CEP without Taylor approximations.
\end{lem}\label{lem:1}
\cmt{
\begin{proof}
	Upon convergence, EP reaches a fixed point such that for $\forall n,k$, 
	\begin{align}
	&\EE_{\hat{p}_i(\btheta_{\setminus m})}[\Phi_m] = \EE_{q(\btheta_{\setminus m})}[\Phi_m] \label{eq:mm1},\\
	&\EE_{\hat{p}_i(\btheta)}\big(\phi(\btheta_{m})\big)  =\EE_{\hat{p}_i(\btheta_{\setminus m})}[h] = \EE_{q(\btheta_{m})}[\phi(\btheta_{m})], \label{eq:mm2}
	\end{align}
	 where the conditional moment $h =   \EE_{\hat{p}_i(\btheta_{m}|\btheta_{\setminus m})} \big(\phi(\btheta_{m})\big)$.  When $h \in \Phi_m$, we have $\EE_{\hat{p}_i(\btheta_{\setminus m})}[h] = \EE_{q(\btheta_{\setminus m})}[h]$ from \eqref{eq:mm1}. Then from \eqref{eq:mm2}, we further have $\EE_{q(\btheta_{\setminus m})}[h] = \EE_{q(\btheta_{m})}[\phi(\btheta_{m})]$, which is the fixed point when CEP converges without Taylor approximations in \eqref{eq:1st}\eqref{eq:2nd}. 
\end{proof}
}
The proof is given in the supplementary material.  While in most cases,  CEP may reach a fixed point different from EP, we found that the quality of the estimated posteriors is close to that of  EP in our experiments. In the difficult cases for standard EP, CEP can easily and rapidly converge to good posterior estimations. 


\begin{algorithm}[htbp]                      
	\caption{Conditional Expectation Propagation (CEP)}          
	\label{alg:cep}                           
	\begin{algorithmic}[1]                    
		\STATE Initialize $q(\btheta) = 1$ and all the messages $\tf_{im}(\btheta_{m})=1$.
		\REPEAT
		\STATE Pick a factor $f_i$ and message $\tf_{im}$, 
		\begin{compactitem}
			\item \textbf{Message deletion:} Calculate the calibrating distribution, $q^{\setminus i}(\btheta_{m}) \propto q(\btheta_{m})/\tf_{im}(\btheta_{m})$. 
			\item \textbf{Projection:} Derive the conditional moment of $\btheta_{m}$ w.r.t $\hat{p}_i({\btheta_{m}|\btheta_{\setminus m}})$ in \eqref{eq:cond-tilt}, and then compute its expectation w.r.t $q(\btheta_{\setminus m})$. If the expectation is intractable, use \eqref{eq:1st} or \eqref{eq:2nd} for approximations. The expected conditional moments are used to construct a new posterior $q^*(\btheta_m)$.
			\item \textbf{Update:} Update the message based on the new posterior: $\tf_{im} \propto q^*(\btheta_m)/q^{\setminus i}(\btheta_{m})$.
		\end{compactitem}
		\UNTIL all the $\tf_{im}(\btheta_{m})$ converge
	\end{algorithmic}
\end{algorithm}
\subsection{ALGORITHM COMPLEXITY}
Given $N$ factors and $M$ dimensional sufficient statistics for $q(\btheta)$, the time and space complexity of CEP using the first-order Taylor approximation and the second-order with diagonal Hessian or covariance matrix are both $\Ocal(NM)$. If we use the full  Hessian and full covariance matrix, the time and space complexity are $\Ocal(NM^2)$. 
\section{CEP FOR BAYESIAN TENSOR DECOMPOSITION AND LOGISTIC REGRESSION}
As a case study, we apply CEP in two popular models, Bayesian tensor decomposition and logistic regression.  
\subsection{BAYESIAN TENSOR DECOMPOSITION}\label{sec:btd}
\vspace{-0.1in}
Denote a $K$-mode tensor by $\Ycal \in \mathbb{R}^{d_1 \times \ldots \times d_K}$, where $d_k$ is the dimension of the $k$-th mode, corresponding to $d_k$ objects (\eg users or items).  The entry value at location $\bi=(i_1,\ldots, i_K)$ is denoted by $y_{\bi}$. We introduce an $R$  dimensional embedding vector $\u^{k}_j$ to represent each object $j$ in mode $k$.  A $d_k \times R$ embedding matrix is formed by stacking all the embedding vectors in mode $k$, $\U^{k} = [{\u_1^{k}}, \ldots, {\u_{d_k}^{k}}]^\top$. Tensor decomposition aims to use these  embeddings $\Ucal = \{\U^{k}\}_k$ to reconstruct the observed entries. 

We use a Bayesian model based on the classical CANDECOMP/PARAFAC (CP) decomposition~\citep{Harshman70parafac} to sample the observed tensor entries. We consider continuous and binary observations. Each embedding vector $\u_j^k$ is first sampled from a Gaussian prior distribution. Given the embeddings, a continuous entry value $y_\bi$ is sampled from 
\begin{align}
p(y_\bi|\Ucal, \tau) = \N(y_\bi |  \1^\top (\u^1_{i_1} \circ \ldots \circ \u^K_{i_K}), \tau^{-1}) \label{eq:ll_c}
\end{align}
where $\1$ is the vector full of ones, $\circ$ the Hadamard (or element-wise) product, and $\tau$ the inverse variance. We assign $\tau$ a Gamma prior, $p(\tau) = \mathrm{Gam}(\tau|a_0, b_0)$.
If $y_\bi$ is binary, it is sampled from 
\begin{align}
p(y_\bi|\Ucal) = \psi\big((2y_\bi -1)\1^\top (\u^1_{i_1} \circ \ldots \circ \u^K_{i_K})\big) \label{eq:ll_b}
\end{align}
where $\psi(\cdot)$ is the cumulative density function (CDF) of the standard Gaussian distribution. 

 To estimate the posterior of the embeddings $\Ucal$ with EP, the major hurdle is to approximate the likelihood of each observation, \eqref{eq:ll_c} or \eqref{eq:ll_b}. 
 Due to the Hadamard product, the log normalizer of the titled distribution is intractable and the moments are tricky to compute. Hence, we turn to CEP.  Note that the prior factors belong to the exponential family and do not need approximations. For the continuous entry, we approximate \eqref{eq:ll_c} by $\tf_\bi(\tau)\prod_k \tf_{\bi}^k(\u_{i_k}^k)$ where $\tf_\bi(\tau)$ is a Gamma term and each $\tf_\bi^k(\u_{i_k}^k)$ a Gaussian term. The approximate posterior $q(\Ucal, \tau)$ and the calibrating distribution $q^{\setminus \bi}(\Ucal, \tau)$ are hence factorized over $\tau$ and  all $\{\u_j^k\}_{j,k}$. To update each message $\tf_\bi^k(\u_{i_k}^k)$, we first derive the conditional moment of $\u_{i_k}^k$ given the other embeddings and $\tau$ fixed. Note that the conditional tilted distribution, 
 \[
 \hat{p}_\bi(\u_{i_k}^k | \u_\bi^{\setminus k}, \tau) \propto q^{\setminus \bi}(\u_{i_k}^k)\N(y_\bi | {\z_\bi^{\setminus k}}^\top \u^k_{i_k}, \tau^{-1}) 
 \]
  is Gaussian. Here $\u_\bi^{\setminus k}$ are all the embedding vectors associated with entry $\bi$ but excluding $\u_{i_k}^k$, $\z_\bi^{\setminus k}$ the Hadmard product of the vectors in $\u_\bi^{\setminus k}$, and $q^{\setminus \bi}(\u_{i_k}^k) = \N(\u^k_{i_k}|\m^k_{i_k}, \S^k_{i_k})$, the calibrating distribution of $\u_{i_k}^k$. We can easily obtain the conditional mean and covariance,
  \begin{align}
  &\mathrm{cov}(\u_{i_k}^k|\u_\bi^{\setminus k}, \tau) = \big[{\S^k_{i_k}}^{-1} + \tau  (\z_\bi^{\setminus k} {\z_\bi^{\setminus k}}^\top)\big]^{-1}, \label{eq:cov} \\  
  &\EE(\u_{i_k}^k|\u_\bi^{\setminus k}, \tau ) = \mathrm{cov}(\u_{i_k}^k|\u_\bi^{\setminus k}, \tau)\big[ {\S^k_{i_k}}^{-1}\m^k_{i_k} + \tau y_\bi \z_\bi^{\setminus k}\big]. \label{eq:mean}
  \end{align}
 \cmt{
  \begin{align}
  &\EE(\u_{i_k}^k) = \bSigma( {\S^k_{i_k}}^{-1}\m^k_{i_k} + \tau y_\bi \z_{i_k}^k), \nonumber \\  
  &\EE(\u_{i_k}^k {\u_{i_k}^k}^\top ) = \bSigma +   \EE(\u_{i_k}^k)  \EE^\top(\u_{i_k}^k)
  \end{align}
  where $\bSigma = \big({\S^k_{i_k}}^{-1} + \tau y_\bi (\z_{i_k}^k {\z_{i_k}^k}^\top)\big)^{-1}$ is the covariance of the conditional titled distribution. 
}
To further obtain their expectations w.r.t $q(\u_\bi^{\setminus k}, \tau)$ so as to update $\tf_\bi^k(\u_{\setminus i_k}^k)$, we can follow \eqref{eq:1st} to take the expectation over their first-order Taylor expansion at the moments of $\u_\bi^{\setminus k}$ and $\tau$. This is trivial --- we replace the terms $\z_\bi^{\setminus k}$, $\z_\bi^{\setminus k}{\z_\bi^{\setminus k}}^\top$ and $\tau$ by their expectations in \eqref{eq:cov}\eqref{eq:mean}. Due to the factorized posterior, we have 
\begin{align}
&\EE_{q}(\z_\bi^{\setminus k}) = \EE_{q}(\u_{i_1}^1) \circ  \ldots \circ \EE_{q}(\u_{i_{k-1}}^{k-1}) \nonumber \\
&\circ \EE_{q}(\u_{i_{k+1}}^{k+1})\circ \ldots \circ \EE_{q}(\u_{i_K}^K), \\
&\EE_{q}(\z_\bi^{\setminus k}{\z_\bi^{\setminus k}}^\top) = \EE_{q}(\u_{i_1}^1{\u_{i_1}^1}^\top) \circ  \ldots \circ \EE_{q}(\u_{i_{k-1}}^{k-1}{\u_{i_{k-1}}^{k-1}}^\top) \nonumber \\
&\circ \EE_{q}(\u_{i_{k+1}}^{k+1}{\u_{i_{k+1}}^{k+1}}^\top)\circ \ldots \circ \EE_{q}(\u_{i_K}^K{\u_{i_K}^K}^\top).
\end{align}
We can use the second-order Taylor approximation as well (see \eqref{eq:2nd}), but our investigation shows that it does not outperform the first-order approach and is slower. 

We now look into how to update message $\tf_\bi(\tau)$. Note that the calibrating distribution is a Gamma distribution, $q^{\setminus \bi}(\tau)= \mathrm{Gam}(\tau|a^{\setminus \bi}, b^{\setminus \bi})$. We first find that the conditional tilted distribution is also a Gamma distribution,  
\begin{align}
&\hat{p}_\bi(\tau|\u_{i_1}^1, \ldots, \u_{i_K}^K)= \mathrm{Gam}(\tau|\hat{a}, \hat{b}) \nonumber \\
&\propto q^{\setminus \bi}(\tau) \N(y_\bi |  \1^\top (\u^1_{i_1} \circ \ldots \circ \u^K_{i_K}), \tau^{-1})  \nonumber
\end{align}
 where $\hat{a} = a^{\setminus \bi} +  \frac{1}{2}$ and $\hat{b} =b^{\setminus \bi} + \frac{1}{2} (y_\bi - \1^\top (\u^1_{i_1} \circ \ldots \circ \u^K_{i_K}))^2$. Next, we take the expectation over $\hat{b}$ w.r.t to the posterior $q(\u_{i_1}^1, \ldots, \u_{i_K}^K)$ ($\hat{a}$ is constant and so $\EE_q(\hat{a}) = \hat{a}$ ). This is analytical and straightforward:  
 \begin{align}
 &\EE_q(\hat{b}) = b^{\setminus \bi} + \frac{1}{2}y_\bi^2 - y_\bi \1^\top \big[\EE_{q}(\u_{i_1}^1) \circ  \ldots \circ \EE_{q}(\u_{i_K}^K)\big] \nonumber \\
 &+\frac{1}{2}\tr\big[\EE_{q}(\u_{i_1}^1{\u_{i_1}^1}^\top)\circ \ldots \circ \EE_{q}(\u_{i_K}^K{\u_{i_K}^K}^\top)\big]. \label{eq:tau}
 \end{align}
 As in Algorithm \ref{alg:cep}, $\EE_q(\hat{a})$ and $\EE_q(\hat{b})$ are then used to build a new Gamma posterior for $\tau$ to update $\tf_\bi(\tau)$.
 
 For binary tensor entries, we use $\prod_k \tf_\bi^k(\u^k_{i_k})$ to approximate the likelihood in \eqref{eq:ll_b}.  The conditional tilted distribution for each message $\tf_\bi^k(\u^k_{i_k})$ is 
 \[
 \hat{p}_\bi(\u_{i_k}^k | \u_\bi^{\setminus k}) \propto q^{\setminus \bi}(\u_{i_k}^k)\psi\big((2y_\bi - 1)  {\z_\bi^{\setminus k}}^\top \u^k_{i_k}\big) .
 \]
 The log normalizer is tractable and so we can derive the conditional moments analytically. They have the same form as the moments required in EP for Bayesian probit regression~\citep{probitEP}. We can then use Taylor approximations \eqref{eq:1st}\eqref{eq:2nd} to compute the expected  conditional moments and to update each $\tf_\bi^k(\u_{i_k}^k)$. The details for both continuous and binary tensors are given in the supplementary material.

\cmt{
The joint probability of our Bayesian tensor decomposition model is 
\begin{align}
&p(\{y_\bi\}_{\bi\in S}, \Ucal, \tau) = \mathrm{Gam}(\tau|a_0, b_0)\prod_{k=1}^K\prod_{s=1}^{d_k} \N(\u_s^k | \m_s^k, v\I) \nonumber \\
&\cdot \prod_{\bi \in S}  \N(y_\bi |  \1^\top (\u^1_{i_1} \circ \ldots \circ \u^K_{i_K}), \tau^{-1}), \label{eq:prob_real}
\end{align}
for continuous data and 
\begin{align}
&p(\{y_\bi\}_{\bi\in S}, \Ucal) = \prod_{k=1}^K\prod_{s=1}^{d_k} \N(\u_s^k | \m_s^k, v\I) \nonumber \\
&\cdot \prod_{\bi \in S}  \phi\big((2y_\bi -1)\1^\top (\u^1_{i_1} \circ \ldots \circ \u^K_{i_K})\big). \label{eq:prob_binary}
\end{align}
for binary data. 

As we can see, the prior factors are already in the exponential family.  To estimate the posterior distribution of the embeddings $\Ucal$ with EP, the trick factors are the likelihoods for each observed entries (see \eqref{eq:ll_c} and \eqref{eq:ll_b}). Due to the production form of the embeddings, the moment matching is intractable. 
}

\subsection{BAYSESIAN LOGISTIC REGRESSION}\label{sec:blr}
Given the observed classification instances $\X =[\x_1, \ldots, \x_n]^\top$ and binary labels  $\y = [y_1, \ldots, y_n]$,  the joint probability of the Bayesian logistic model is 
\begin{align}
p(\y, \w|\X) = p(\w) \prod_{i=1}^n 1/\big(1  + \exp(-(2y_i - 1)\w^\top \x_i)\big),  \nonumber 
\end{align}
where the prior $p(\w)$ is usually chosen as a Gaussian distribution.
The moment matching in regard to the logistic likelihood is intractable, and there is a smart approximation~\citep{gelman2013bayesian}. Here, we consider CEP instead. We choose fully factorized messages $\prod_m\tf_{im}(w_m)$ to approximate each logistic likelihood, where each message $\tf_{im}(w_m)$ is Gaussian and so are $q(w_m)$ and $q^{\setminus i}(w_m)$. The conditional tilted distribution is 
\[
\hat{p}_i(w_m|\w_{\setminus m}) \propto q^{\setminus i}(w_m) g_{im}(w_m|\w_{\setminus m}), 
\]
where $\w_{\setminus m}$ are $\w$ excluding $w_m$, and $g_{im}(w_m|\w_{\setminus m}) = 1/\big(1+\exp(-(2y_i  - 1)w_mx_{im} - (2y_i - 1)\w_{\setminus m}^\top \x_{i\setminus m})\big)$. Here $x_{im}$ is the $m$-th element of $\x$ and $\x_{i\setminus m}$ the remaining elements. 
The moments of $w_m$ are still intractable. Nevertheless, because the conditional distribution is for a single variable, we can approximate the moments by a quadrature formula. Since $q^{\setminus i}(w_m)$ is Gaussian, we can use Gauss-Hermite quadrature. Given quadrature nodes and weights $\{(\gamma_j, \alpha_j)\}_j$, the conditional moments can be represented by 
\begin{align}
\EE(w_m|\w_{\setminus m}) \approx \frac{\sum_j \alpha_j \gamma_j g_{im}(\gamma_j|\w_{\setminus m})}{\sum_j \alpha_j g_{im}(\gamma_j|\w_{\setminus m})}, \\
\EE(w_m^2|\w_{\setminus m}) \approx \frac{\sum_j \alpha_j \gamma^2_j g_{im}(\gamma_j|\w_{\setminus m})}{\sum_j \alpha_j g_{im}(\gamma_j|\w_{\setminus m})}.
\end{align}
Note that the quadrature nodes are determined by $q^{\setminus i}(w_m)$; both the nodes and weights are constant to $\w_{\setminus m}$.

To update $\tf_{im}$,  we compute the expectation of the conditional moments w.r.t $q(\w_{\setminus m})$. To this end, we can use their first-order or second-order Taylor approximations and then take expectation as shown in \eqref{eq:1st} and \eqref{eq:2nd}. Both are straightforward to derive and efficient to calculate. The details are provided in the supplementary material.


\vspace{-0.05in}
\section{RELATED WORK}
\vspace{-0.05in}
Expectation propagation (EP)~\citep{minka2001expectation} is a deterministic approximate inference algorithm that unifies assumed-density-filtering (ADF) ~\citep{maybeck1982stochastic,lauritzen1992propagation,boyen1998tractable} and loopy belief propagation~\citep{murphy1999loopy}. It iteratively minimizes (local) KL divergence via moment matching to update factor approximations, which can be viewed as message passing and updates in graphical model representations. EP updates are essentially fixed point iterations.  Although the convergence is not guaranteed,  EP often converges fast in practice and produces accurate posterior estimations. EP can be further extended to power EP~\citep{minka2004power} that minimizes the $\alpha$-divergence during the message update.\cmt{, which is called power EP~\citep{minka2004power}.} Another important variant is stochastic EP (SEP)~\citep{li2015stochastic}, which only stores  and updates a single or a few  approximate factors and hence can largely reduce the memory cost. 

Using EP can be troublesome when the moment matching is intractable. While importance sampling can solve this problem in principle, it needs massive samples to obtain a reliable estimation and is very expensive. To address this issue, 
\citet{eskin2004laplace} developed Laplace propagation (LP) to find the Laplace approximation to the tilted distribution. While getting rid of sampling, LP requires repeated numerical optimizations during the message updates and can still be costly. Recently, several excellent works propose to use machine learning models to predict the moments~\citep{heess2013learning,eslami2014just,jitkrittum2015kernel}.The training examples are collected by running importance sampling in the first a few EP iterations. \citet{heess2013learning} used neural-networks, \citet{eslami2014just} random forests and \citet{jitkrittum2015kernel} Gaussian processes with random features.  In \citep{jitkrittum2015kernel, eslami2014just}, the predictive variance or an uncertainty score is used to monitor the quality of the predicted moments -- if it exceeds a threshold,  importance sampling is used again to improve the estimation and to replenish the training data. Despite the promising performance, these methods have to run a lot of importance sampling to collect sufficient training examples, which might still be expensive. Moreover, extra efforts need to be made to design models, extract features and choose hyper-parameters. There are also concerns in the generalizability to new data points; the learned model only specializes on one type of factors, and usually cannot generalize to other types. To address these practical barriers, we propose CEP that bypasses the tricky moment matching, instead seeks for easy and analytical conditional moments and then takes the expectation over the conditional moments. The expectation can be efficiently calculated through Taylor approximations. In this way, CEP avoids to manually design approximations or to run costly importance sampling and numerical optimizations. It can still conduct efficient and analytical message refinements. We can immediately adapt CEP to a stochastic or streaming version, like SEP and ADF. 


\vspace{-0.1in}
\section{EXPERIMENT}
\vspace{-0.1in}
\subsection{BAYESIAN PROBIT AND LOGISTIC REGRESSION}
\vspace{-0.1in}
We first examined our approach on two classical models, Bayesian logistic regression (BLR) and probit regression (BPR). EP has analytical forms for moment matching in BPR. While in BLR the moments are intractable due to the logistic function, we can develop accurate approximations based on quadrature rules~\citep{gelman2013bayesian}. 
\begin{figure*}
	\centering
	\setlength\tabcolsep{0.1pt}
	\begin{tabular}[c]{cccc}
		\begin{subfigure}[t]{0.25\textwidth}
			\centering
			\includegraphics[width=\textwidth]{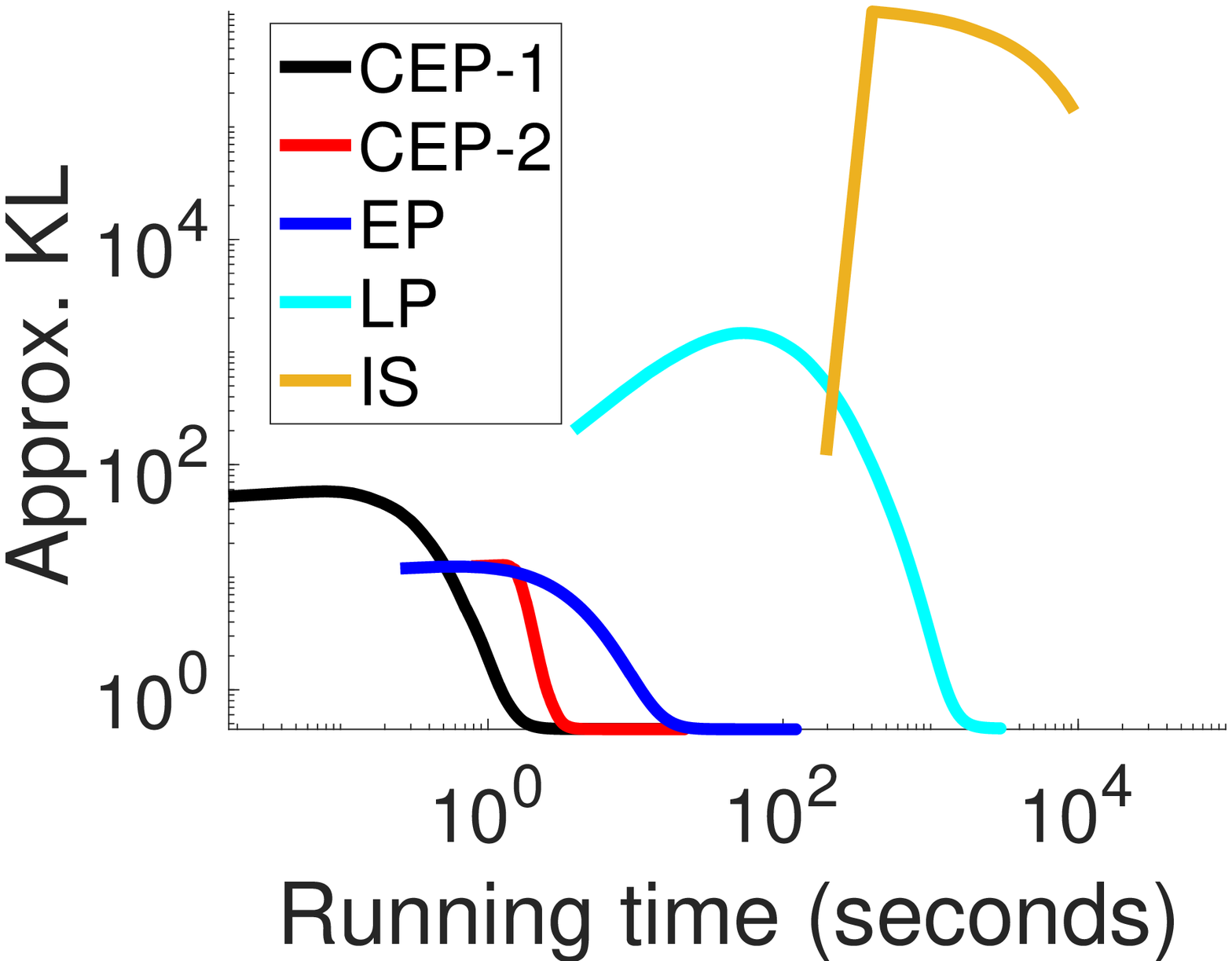}
			\caption{Simu1 in BLR}
		\end{subfigure}
		&
		\begin{subfigure}[t]{0.25\textwidth}
			\centering
			\includegraphics[width=\textwidth]{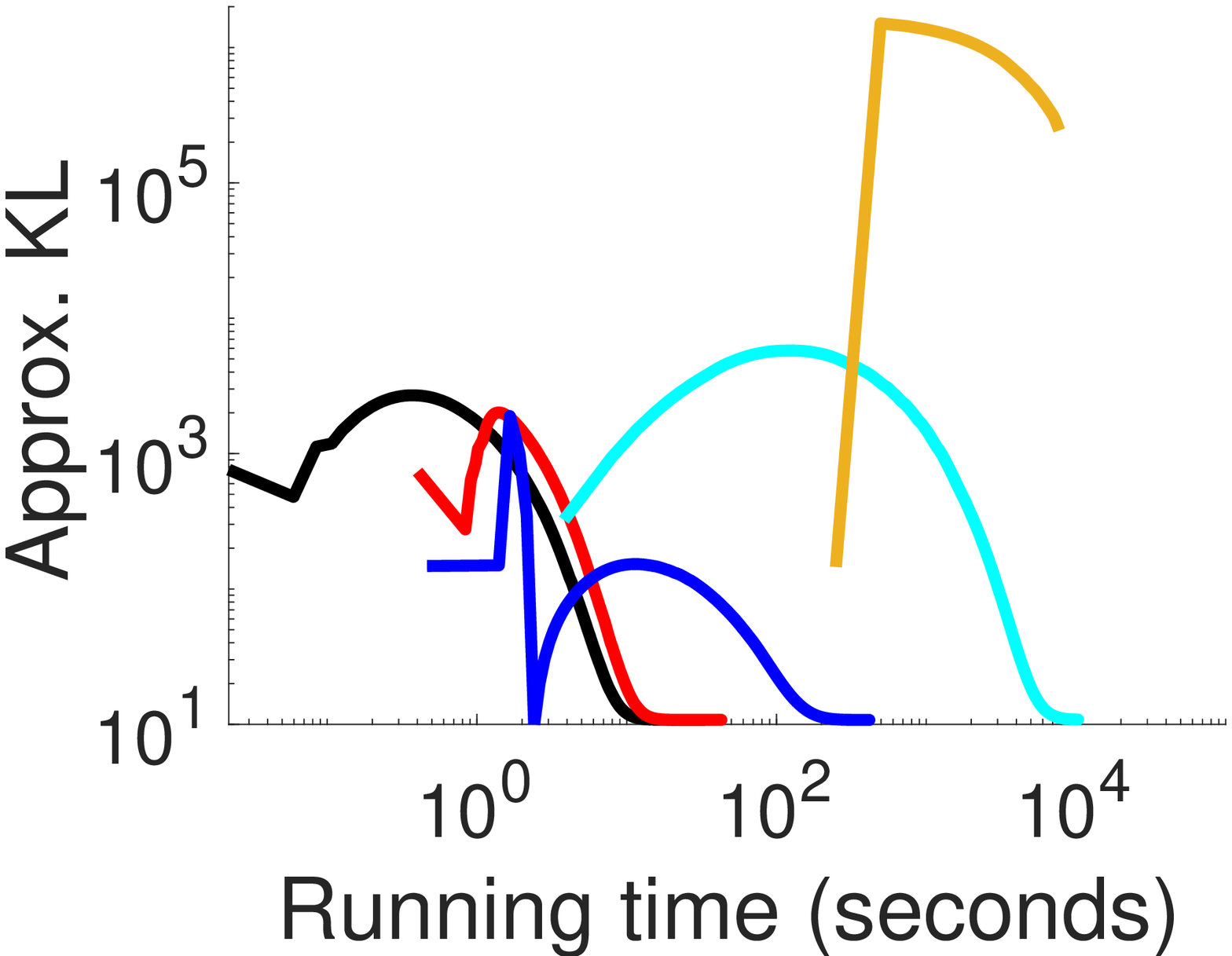}
			\caption{Simu2 in BLR}
		\end{subfigure}  
		&
				\begin{subfigure}[t]{0.25\textwidth}
			\centering
			\includegraphics[width=\textwidth]{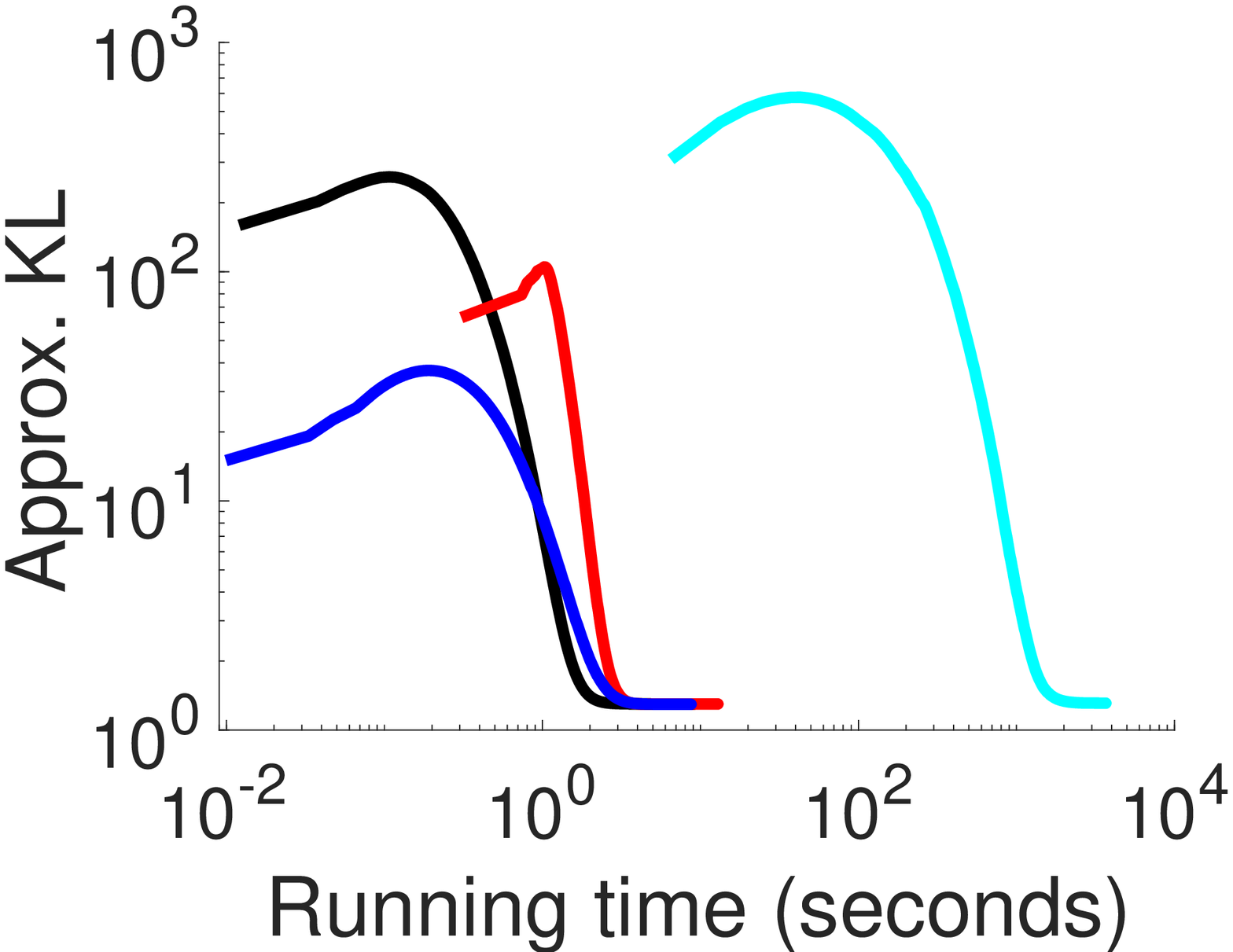}
			\caption{Simu1 in BPR}
		\end{subfigure}
		&
		\begin{subfigure}[t]{0.25\textwidth}
			\centering
			\includegraphics[width=\textwidth]{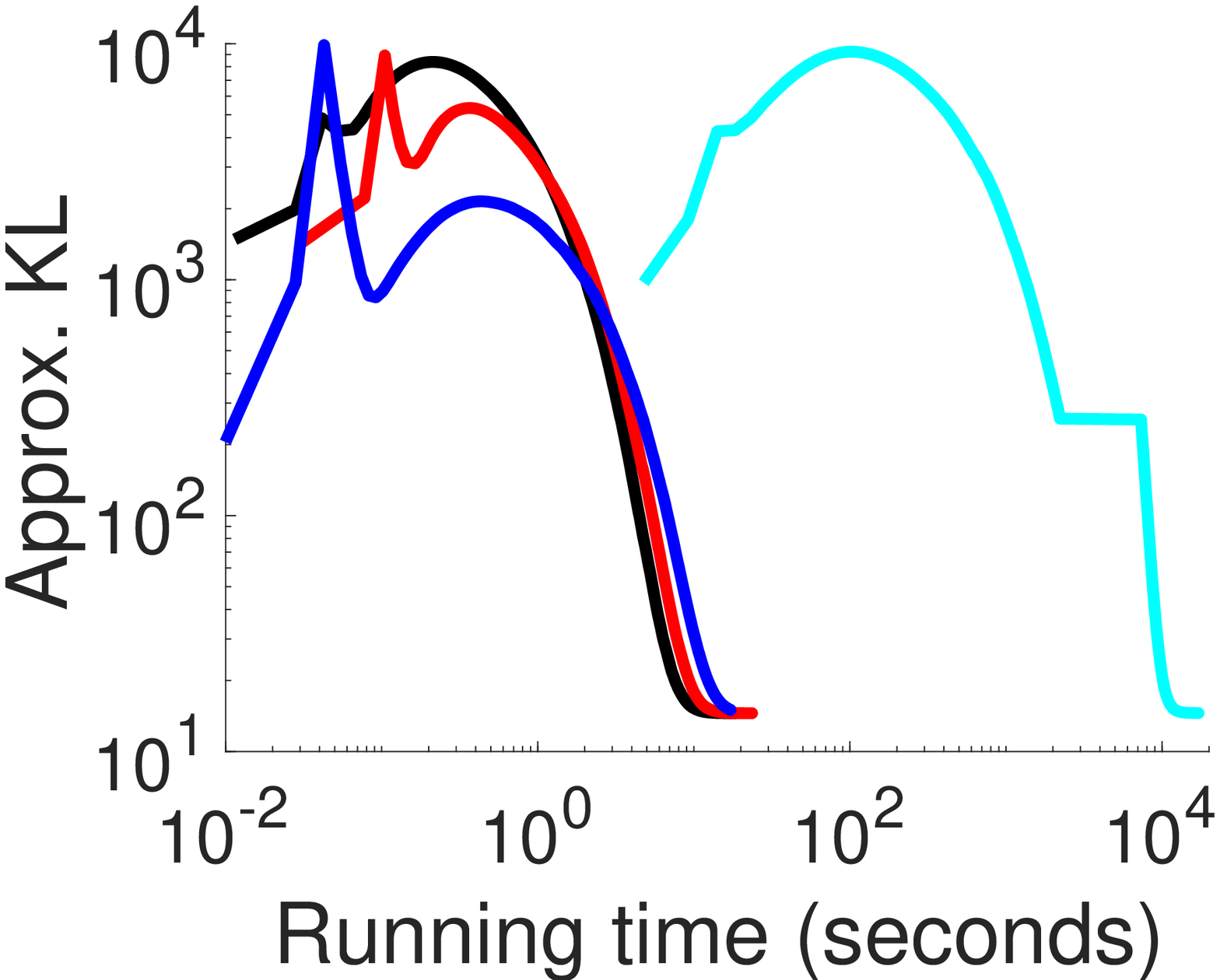}
			\caption{Simu2 in BPR}
		\end{subfigure}
	\end{tabular}
	\vspace{-0.1in}
	\caption{Running time \textit{vs.} approximate KL divergence. BPR and BLR are short for Bayesian probit regression and Bayesian logistic regression, respectively.} 
	\label{fig:simu}
	\vspace{-0.2in}
\end{figure*}

\textbf{Methods.} 
As in~\citep{graepel2010web}, we used fully factorized messages. This is important for large-scale applications. We implemented two versions of the proposed algorithm: CEP-1 that uses the first-order Taylor approximation (see \eqref{eq:1st}) and CEP-2 the second-order (see \eqref{eq:2nd}).  Note that due to the fully factorized approximation, CEP-2 is equivalent to using the diagonal Hessian and covariance in the Taylor approximation. In BLR, we used the Gauss-Hermite quadrature to represent the conditional moments with $9$ nodes (see Section \ref{sec:blr}). We compared with EP, Laplace propagation (LP),  EP with importance sampling (IS) for  moment matching, and variational Bayes(VB).  
For a fair comparison, these methods employed fully factorized posteriors as well.
In addition, we compared with the kernel based just-in-time (KJIT) learning to predict the moments~\citep{jitkrittum2015kernel} for EP updates. KJIT uses IS to dynamically generate training examples and update the predictor. 
 For LP, we ran L-BFGS to find the mode and constructed the Laplace approximation in each message update. The maximum number of iterations was set to $100$. For factorized EP in BLR, we followed~\citep{gelman2013bayesian} to develop a two dimensional Gauss-Hermite quadrature to compute the moments (see Section 2.2.1 \cmt{\ref{sec:sup:ep} }of the supplementary material for details). Note that the original approach in~\citep{gelman2013bayesian} is inappropriate for factorized EP, resulting in much worse performance. For IS and KJIT, we generated $500K$ samples to estimate each moment. Note that we only tested IS and KJIT for BLR, where the exact moments are intractable. 
 We used a flat Gaussian, $\N(\cdot|0, 10^6)$, as the initialization for all the messages in EP, CEP-1, CEP-2, LP and IS. For VB, the initial posterior of each variable was set to the prior.  We used the original implementation of KJIT with C\# under Infer.Net framework (\url{https://github.com/wittawatj/kernel-ep}), which employs a joint Gaussian  term to approximate each likelihood (rather than a factorized one). All the other methods were implemented with MATLAB 2017. We ran all the algorithms on a single Linux server with Intel(R) i7 CPUs and 24GB memory.

\textbf{Synthetic datasets.} To examine the inference quality of our approach, we followed \citep{li2015stochastic} to simulate two datasets for BLR and BPR, respectively.  Each dataset include $10,000$ samples and $4$ classification weights. Each weight was sampled from the standard normal prior $\N(\cdot|0,1)$. In the first dataset for each model, the feature values of each instance $\x_n$ were independently sampled from a single Gaussian distribution $\N(\cdot|0,1)$ while in the second dataset from a mixture of Gaussian distributions with $5$ components,  $\frac{1}{5}\big(N(\cdot |-2,\frac{1}{2}) + N(\cdot |-1,\frac{1}{2}) + N(\cdot |0,\frac{1}{2}) + N(\cdot |1,\frac{1}{2}) + N(\cdot |2,\frac{1}{2})\big)$. The label $y_n$ was sampled from the generative model. Hence, the second dataset for each model is  more heterogeneous. To evaluate the inference quality, we ran No-U-Turn (NUTS)~\citep{hoffman2014no} sampler for $100K$ iterations as the burn-in time and then drew $50K$ posterior samples, from which we estimated a Gaussian distribution as the golden standard.  
We then ran CEP-1, CEP-2, EP, LP, IS and KJIT to obtain the posterior estimations of the classification weights. We computed the KL divergence between the golden standard and the posteriors estimated by each method.  Note that we did not compare with VB because VB minimizes the KL divergence in a reverse direction and this metric may not be fair to VB. 
Although taking much longer running time than CEP-\{1,2\}, EP and LP, KJIT still obtains much larger approximate KL divergence, implying much worse inference quality or even a failure.  The running time and the KL divergence of KJIT are \{$9595.4$, $10150.3$\} seconds and  \{$14.7$, $1.4e13$\} on the two simulation datasets for BLR. 
We show for all the other methods how the KL divergence varies along with running time in Fig. \ref{fig:simu}. As we can see, similar to KJIT, IS spent the most running time but resulted in the worst performance (Fig. \ref{fig:simu}a and b). This might be due to the unstable accuracy of the estimated moments, even with $500K$ samples. In both BPR and BLR, our algorithms, CEP-1 and CEP-2, obtain almost the same approximate KL divergence with EP, implying the same inference quality. Although LP ends up with  a similar KL divergence as well, it is far slower than CEP-1, CEP-2 and EP. CEP-1 is slightly faster than CEP-2, because CEP-2 uses the second order Taylor approximation and needs extra computation. 
 While the running time of EP, CEP-1 and CEP-2 are close in BPR, EP is slower than both CEP-1 and CEP-2 in BLR. The reason might be that EP has to use a two-dimensional Gauss-Hermite quadrature for moment matching, while CEP-1 and CEP-2 just one dimensional quadrature. 
 \cmt{
\begin{figure}[t]
	\centering
	\begin{subfigure}{0.35\textwidth}
		\includegraphics[width=\textwidth]{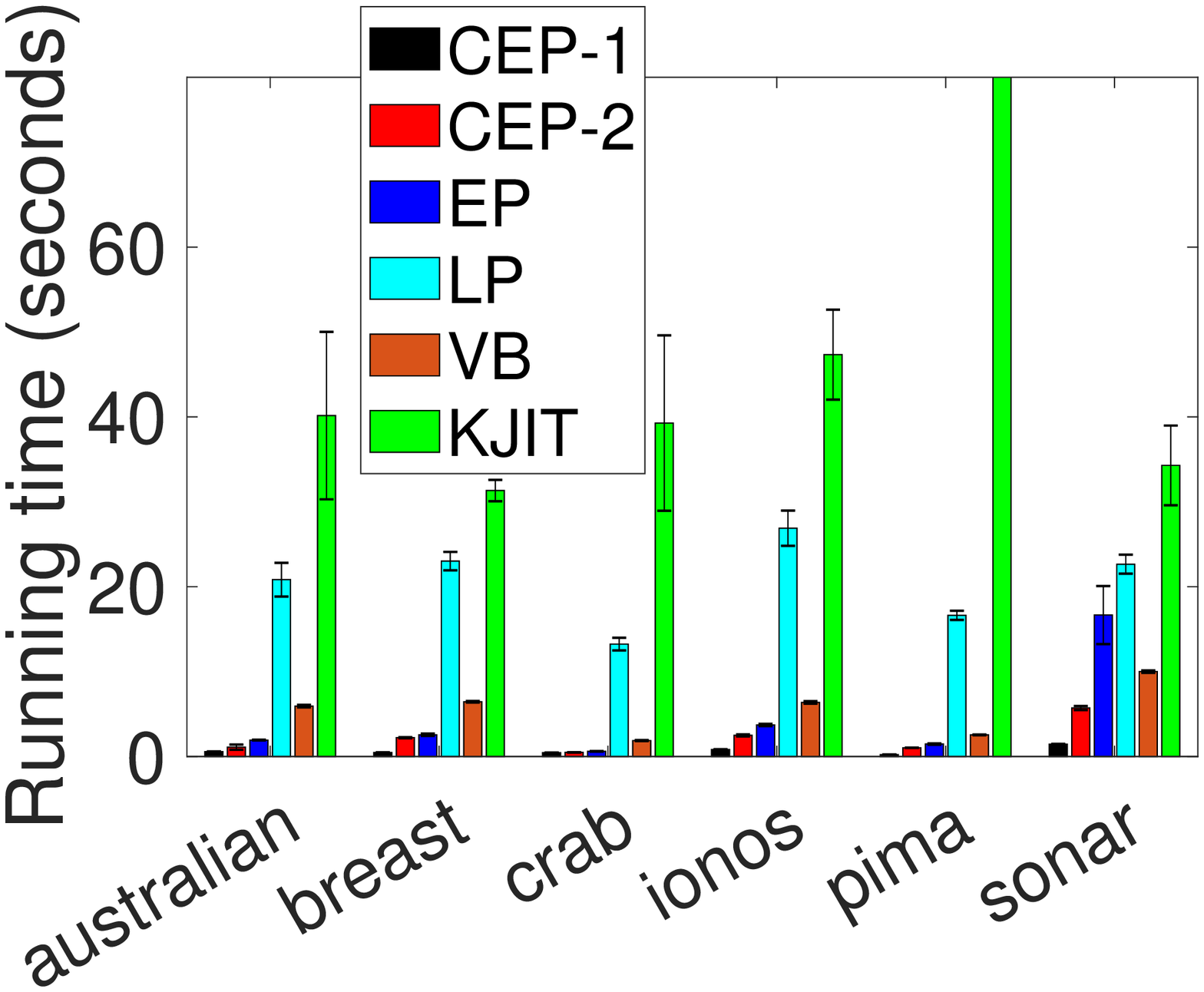}
		\caption{Bayesian logistic regression}
	\end{subfigure}
	\begin{subfigure}{0.35\textwidth}
		\includegraphics[width=\textwidth]{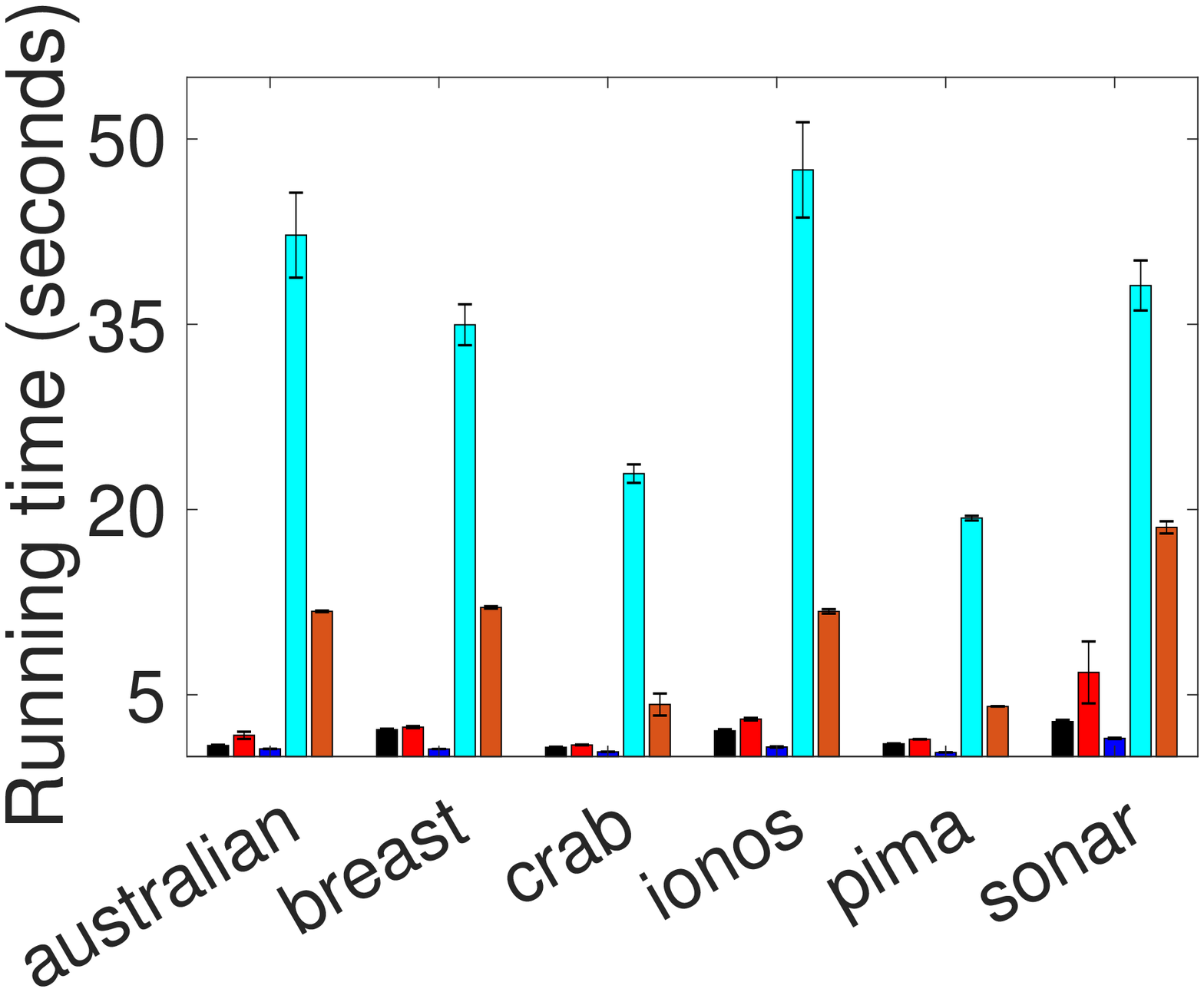}
		\caption{Bayesian probit regression}
	\end{subfigure}
   \vspace{-0.1in}
	\caption{Average running time on six real datasets.}
	\label{fig:time}
	\vspace{-0.2in}
\end{figure}
}
\begin{figure}
	\vspace{-0.05in}
	\centering
	\setlength\tabcolsep{0.1pt}
	\begin{tabular}[c]{cc}
		\begin{subfigure}[t]{0.25\textwidth}
			\centering
			\includegraphics[width=\textwidth]{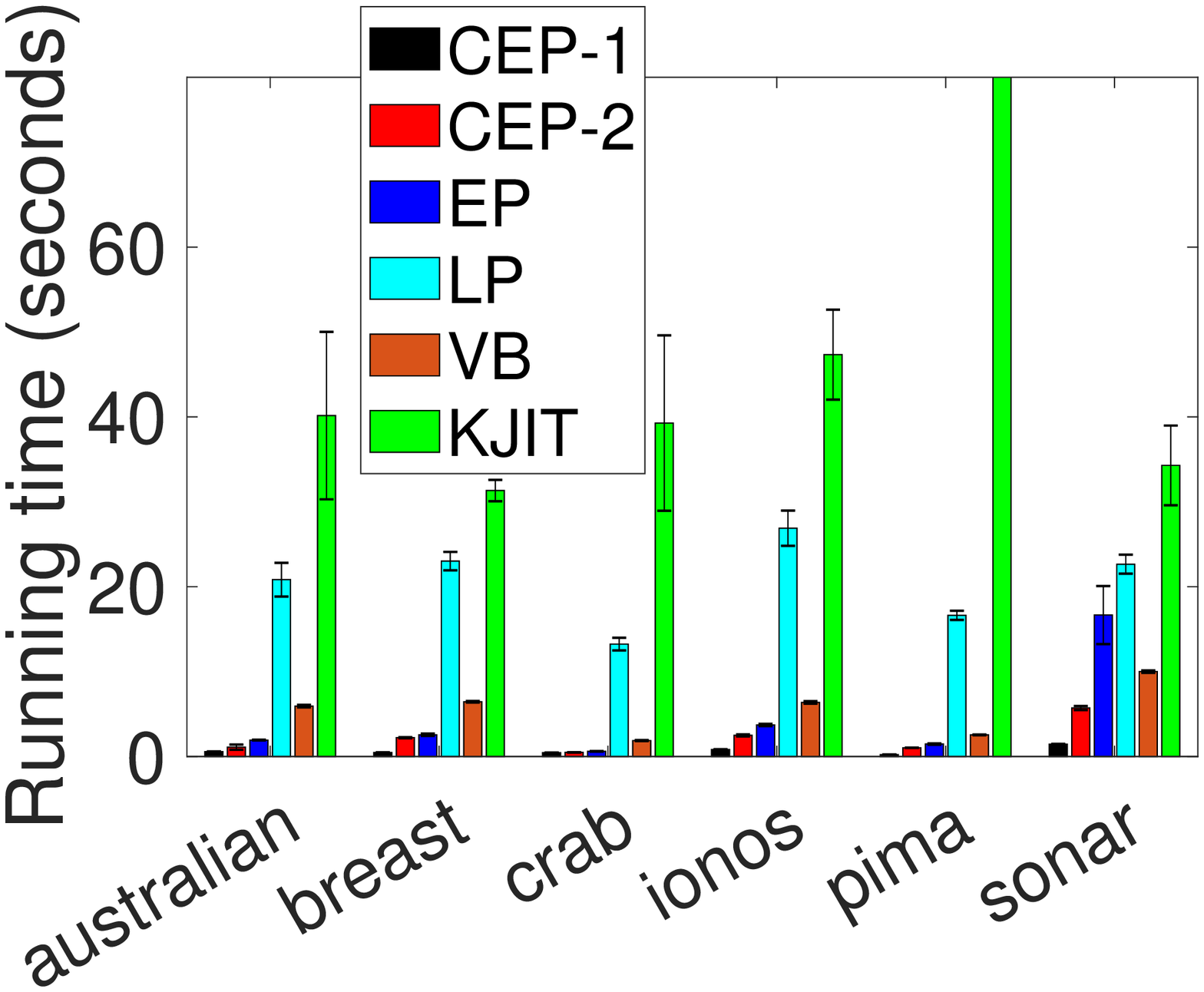}
			\caption{Bayesian logistic regression}
		\end{subfigure}
		&
		\begin{subfigure}[t]{0.25\textwidth}
			\centering
			\includegraphics[width=\textwidth]{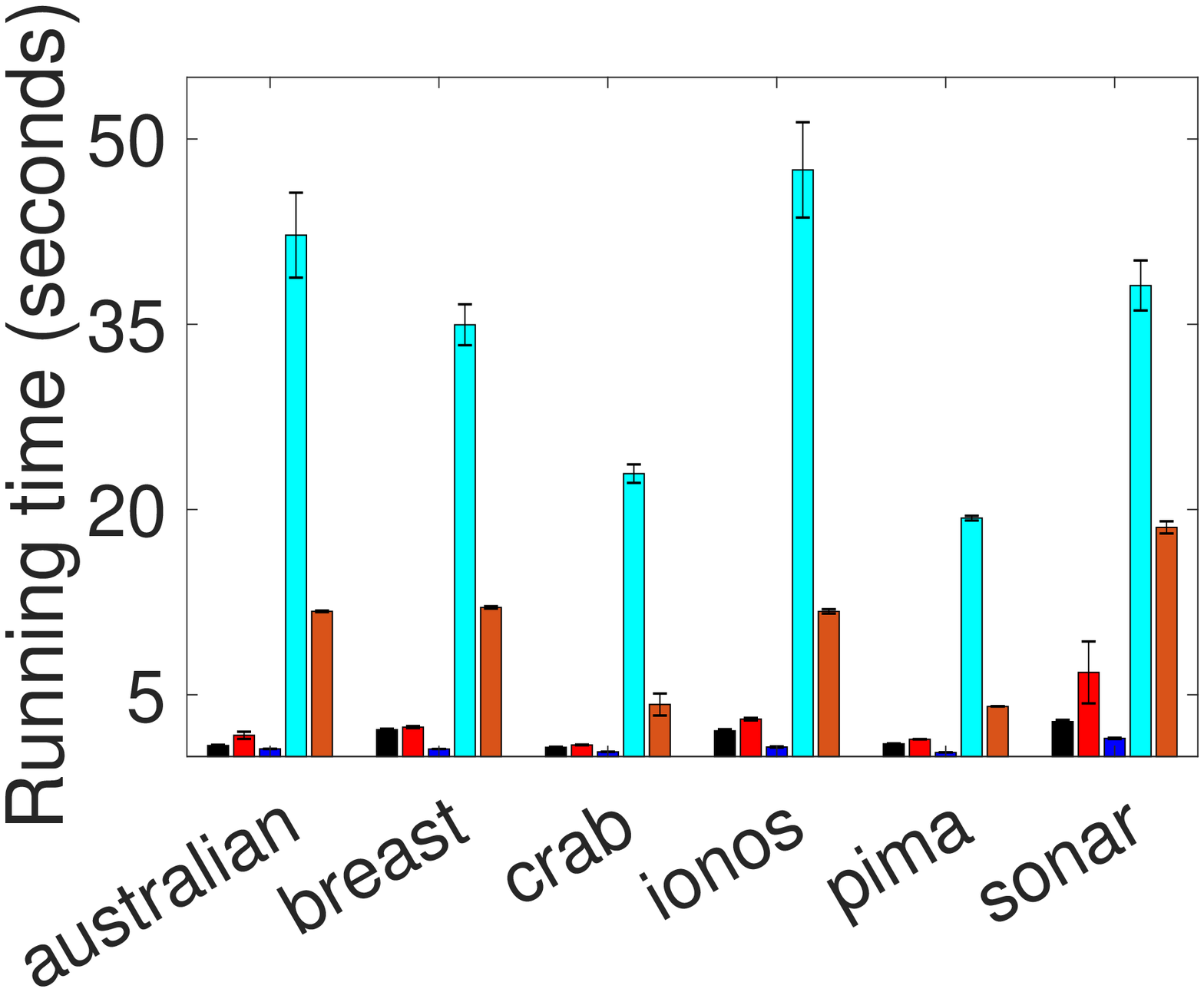}
			\caption{Bayesian probit regression}
		\end{subfigure}  
	\end{tabular}
	\vspace{-0.1in}
	\caption{Average running time on six real datasets.}
	\label{fig:time}
	\vspace{-0.25in}
\end{figure}

\textbf{Real datasets.} To further examine the predictive performance and computational efficiency, we tested all the methods on six real-world datasets from UCI machine learning repository{(\url{https://archive.ics.uci.edu/ml/index.php})}, \textit{australian}, \textit{breast}, \textit{crab},\textit{ionos},  \textit{pima} and \textit{sonar}. We randomly split each dataset into a half for training and the other half for testing. We split the dataset for five times and reported the average test log likelihood, area under ROC curve (AUC), and  running time of each method. The test log likelihoods are summarized in Table \ref{tb:ll}. Due to the space limit, the AUC results are provided in the supplementary material. The running time are shown in Fig. \ref{fig:time}.

From Table \ref{tb:ll}a and b, we can see that CEP-1 and CEP-2 nearly always obtain average test log-likelihoods (a bit) larger than or close to that of EP. It demonstrates that our approach can have the same predictive performance as EP. In most cases, CEP-2 is slightly better than CEP-1, and this might be due to the usage of the second-order Taylor approximations. LP achieves a close prediction accuracy to EP as well, but it is much slower (see Fig. \ref{fig:time}).  VB performs best in \textit{crab} dataset, but is suboptimal in the other datasets. KJIT, however, performs the worst in most cases, especially on \textit{pima} and \textit{sonar}. In fact, KJIT's performance is quite unstable. It often results in an exceedingly small log-likelihood, implying a complete failure. We have to run several times to guarantee reasonable accuracy. 


Fig. \ref{fig:time} shows the average running time  of each method. 
As we can see, CEP-1, CEP-2 and EP are much faster than LP, which is consistent with the results from the simulation (see Fig. \ref{fig:simu}). Note that KJIT is much slower than LP, especially for \textit{pima}, where KJIT's average running time is 904.626 seconds. CEP-1 and CEP-2 are also significantly faster than  VB. In BPR, the speeds of CEP-1 and CEP-2 are close to that of EP, while in BLR, EP is slower, especially on \textit{sonar}, which might be again due to the two dimensional quadrature for moment matching. The results confirm the computational advantage of our approach.  

\begin{table*}
	
\begin{subtable}{\textwidth}
	\centering
	\small
	\setlength\tabcolsep{1.2pt}
	\begin{tabular}[c]{ccccccc}
		\hline\hline
		Dataset & CEP-1 & CEP-2 & LP &  EP & VB & KJIT\\
		\hline 
		australian & $\mathbf{-0.449 \pm 	0.012}$ & $-0.450 \pm 0.015$ & $-0.451 \pm	0.012$ & $-0.450 \pm 	0.015$ & ${-0.449 \pm 0.015}$  & $-0.451 \pm 	0.013$\\
		breast & $-0.579 \pm	0.020$ & $\mathbf{-0.565 \pm 	0.020}$ & $-0.576 \pm	0.018$ & $-0.585 \pm 	0.024$ & $-0.587\pm	0.023$  & $-0.576 \pm	0.017$\\
		crab & $-0.316 \pm	0.004$ & $-0.315 \pm	0.003$ &
		$-0.349 \pm	0.012$ & $-0.313 \pm	0.003$ & 
		$\mathbf{-0.277 \pm	0.004}$ & $-0.327 \pm	0.004$ \\
		ionos & $-0.316 \pm	0.022$ & $\mathbf{-0.302 \pm 0.023}$ & $-0.316 \pm 0.022$ &  $-0.309 \pm	0.025$ & 
		$-0.332 \pm	0.032$  & $-0.339 \pm	0.018$\\
		pima & $-0.541 \pm	0.007$ & 
		$\mathbf{-0.540 \pm	0.006}$ &
		$-0.541 \pm  0.006$ & 
		$-0.541 \pm	0.007$ &
		$-0.542 \pm	0.007$ & $-0.608 \pm	0.014$ \\
		sonar & $-0.522 \pm	0.017$ &
		$\mathbf{-0.513	\pm 0.027}$ &
		$-0.519 \pm	0.016$ &  $-0.531 \pm	0.026$ &
		$-0.578 \pm	0.026$ & $-1.085 \pm	 0.053$\\ 
		\hline 
	\end{tabular}
	\caption{Bayesian logistic regression} 
\end{subtable}
\begin{subtable}{\textwidth}
     \centering
	\small
	\begin{tabular}[c]{cccccc}
		\hline\hline
		Dataset & CEP-1 & CEP-2 & LP &  EP & VB \\
		\hline 
		australian & $\mathbf{-0.428 \pm 	0.012}$ & $-0.431 \pm 0.015$ & $\mathbf{-0.428 \pm	0.012}$ & $-0.435 \pm 	0.015$ & $-0.441 \pm 0.017$ \\		
		breast & $\mathbf{-0.583 \pm	0.018}$ & $-0.592 \pm  0.023$ & $-0.594 \pm	0.023$ & $-0.615 \pm  0.032$ & $-0.621\pm	0.035$ \\
		crab & $-0.228 \pm	0.009$ & ${-0.226 \pm	0.009}$ &
		$-0.231 \pm	0.009$ &		
		$-0.250 \pm	0.008$ &
		$\mathbf{-0.197 \pm	0.010}$ \\
		ionos & $-0.286 \pm	0.012$ & $\mathbf{-0.277 \pm	0.015}$ & $-0.319 \pm 0.009$ &
		$-0.307 \pm	0.010$ & $-0.458 \pm 	0.036$ \\
		pima & $\mathbf{-0.553 \pm	0.009}$ & 
		$-0.554 \pm	0.009$ & $\mathbf{-0.553\pm	0.009}$ &
		$-0.554 \pm	0.009$ &
		$-0.557 \pm	0.010$ \\
		sonar & $-0.533 \pm	0.032$ & $\mathbf{-0.528 \pm	0.032}$ & $-0.579 \pm	0.030$ & $-0.553 \pm	0.039$ & $-0.891 \pm	0.070$\\
		\hline 
	\end{tabular}
	\caption{Bayesian probit regression.} 
	\end{subtable}
\vspace{-0.15in}
\caption{Average test log-likelihoods on six real datasets.}
\label{tb:ll}
\vspace{-0.1in}
\end{table*}

\begin{figure*}
	\centering
	\setlength\tabcolsep{0.1pt}
	\begin{tabular}[c]{cccc}
		\begin{subfigure}[t]{0.25\textwidth}
			\centering
			\includegraphics[width=\textwidth]{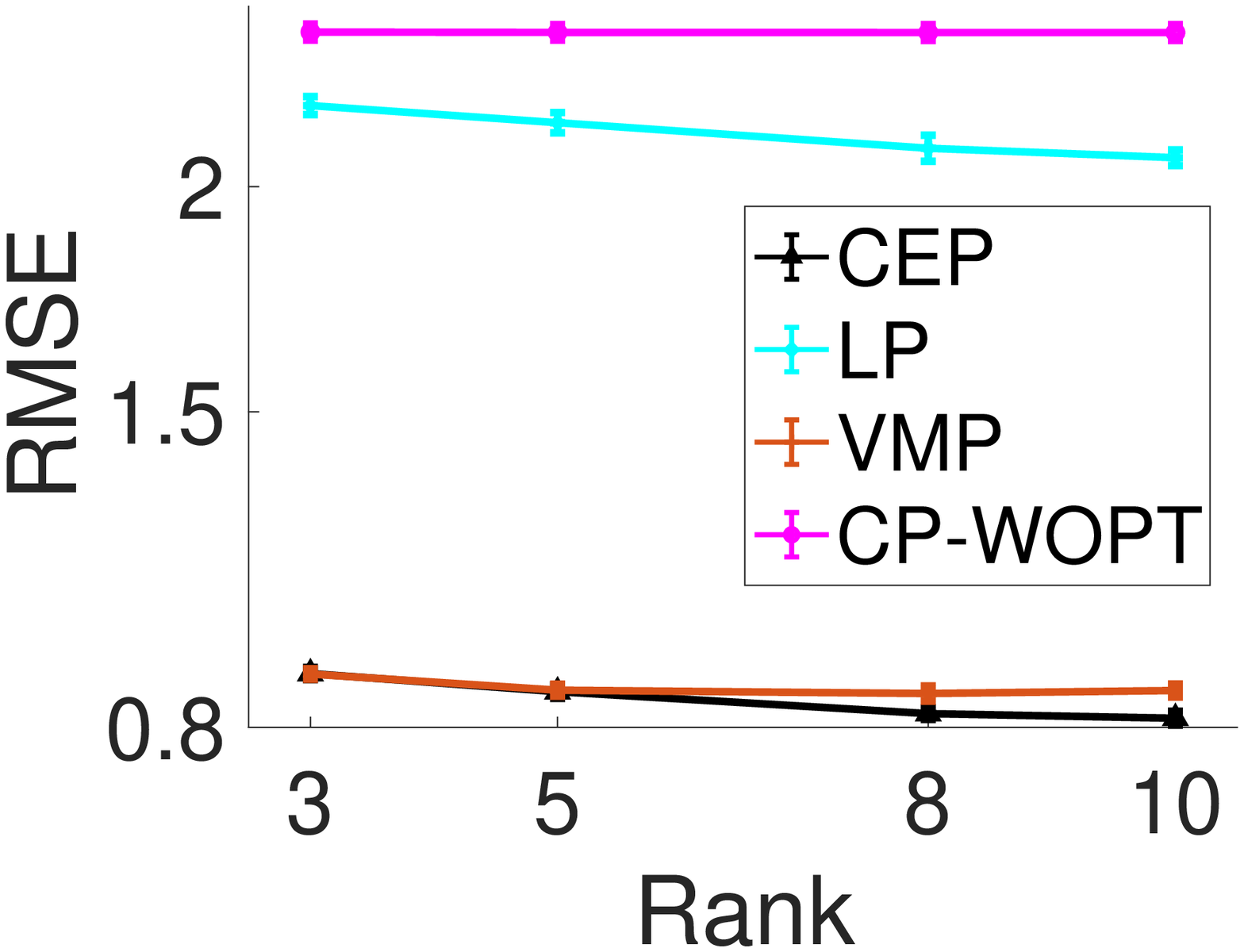}
			\caption{\textit{Alog}}
		\end{subfigure}
		&
		\begin{subfigure}[t]{0.25\textwidth}
			\centering
			\includegraphics[width=\textwidth]{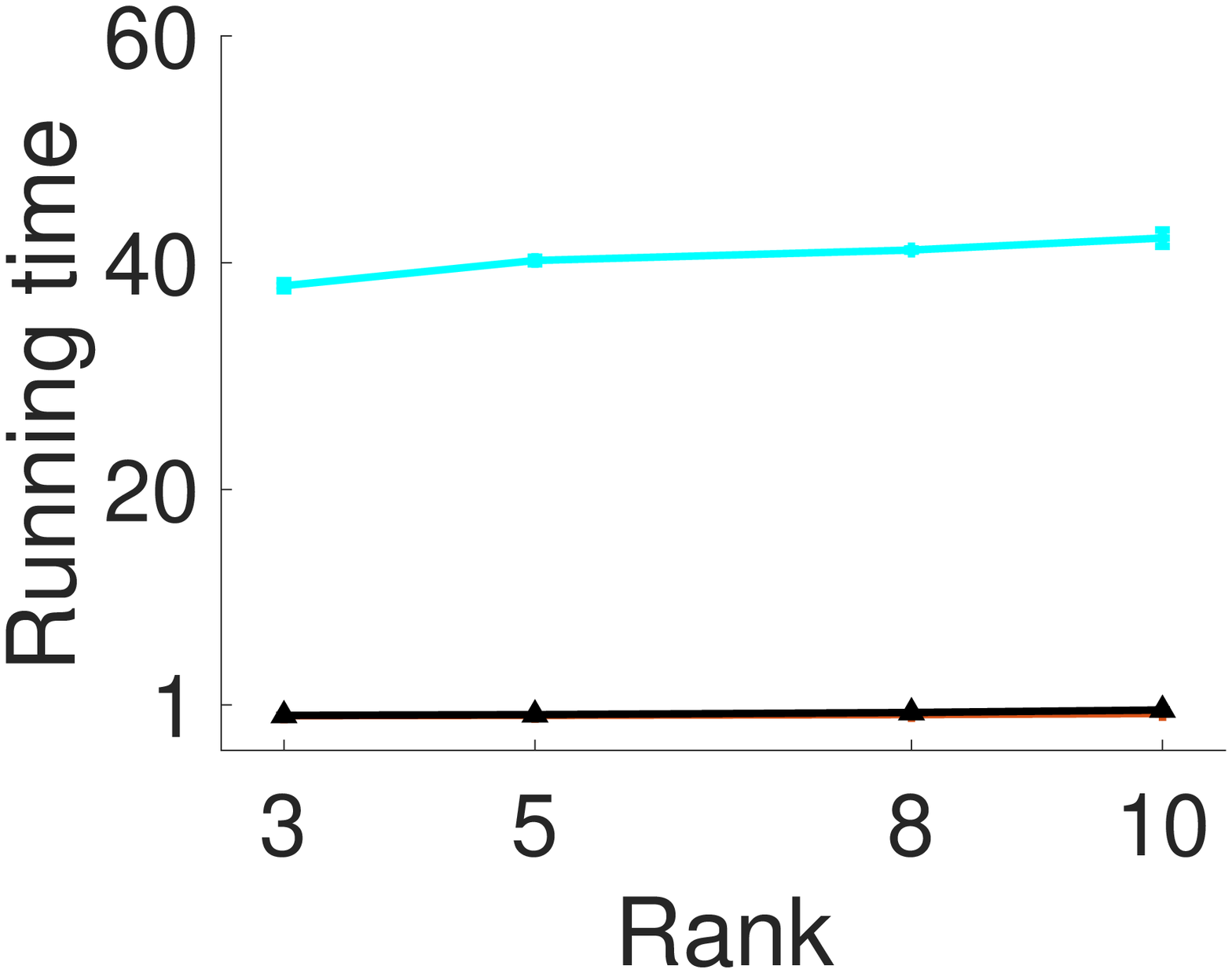}
			\caption{\textit{Alog} running time}
		\end{subfigure}
		&
		\begin{subfigure}[t]{0.25\textwidth}
			\centering
			\includegraphics[width=\textwidth]{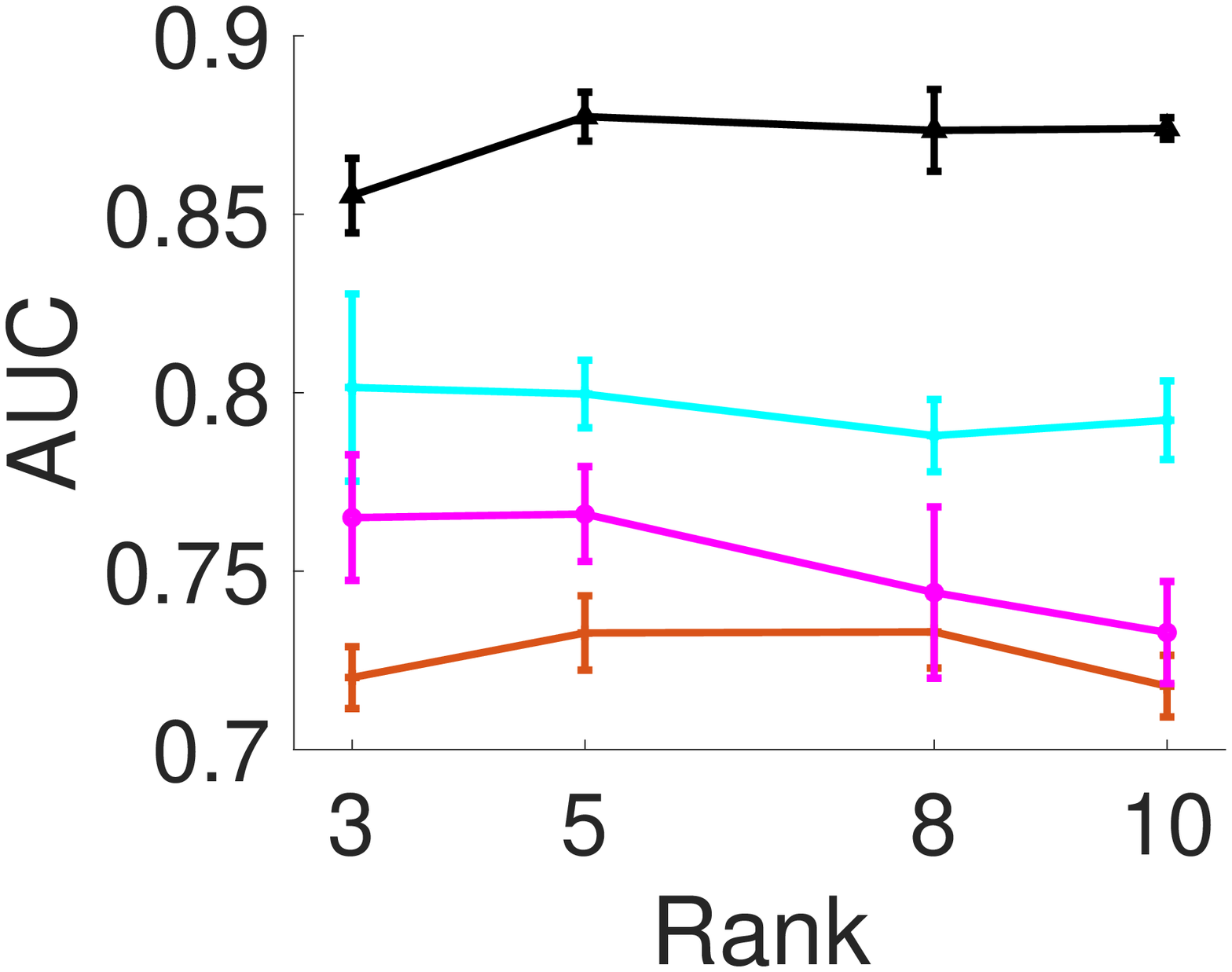}
			\caption{\textit{Enron}}
		\end{subfigure}
		&
		\begin{subfigure}[t]{0.25\textwidth}
			\centering
			\includegraphics[width=\textwidth]{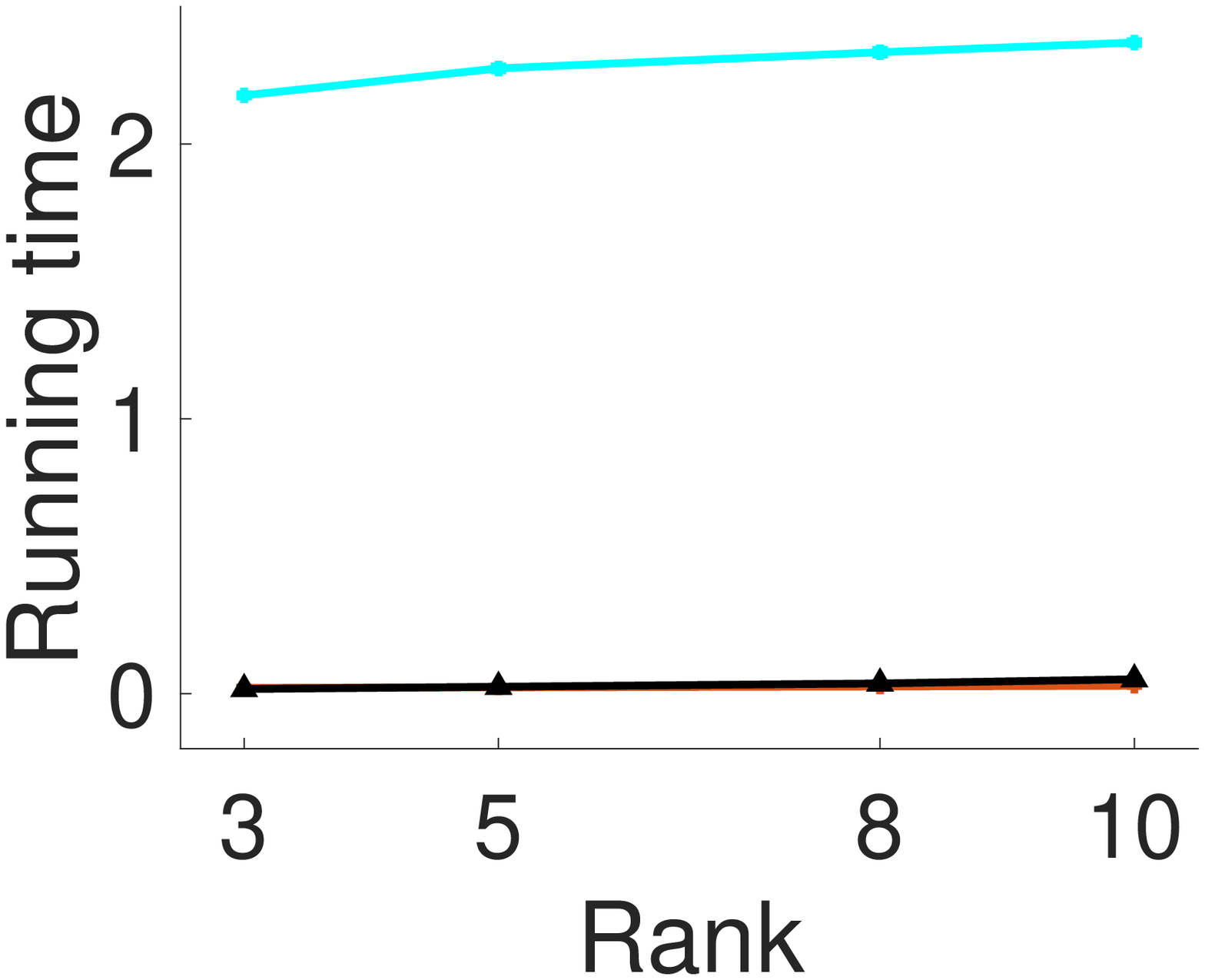}
			\caption{\textit{Enron} running time}
		\end{subfigure}  
	\end{tabular}
	\vspace{-0.1in}
	\caption{Prediction accuracy and per-iteration running time of all the methods for tensor decomposition. Note that the running time of VMP and CEP are close.} 
	\label{fig:tf}
	\vspace{-0.1in}
\end{figure*}

\begin{figure}
	\vspace{-0.2in}
	\centering
	\setlength\tabcolsep{0.1pt}
	\begin{tabular}[c]{cc}
		\begin{subfigure}[t]{0.25\textwidth}
			\centering
			\includegraphics[width=\textwidth]{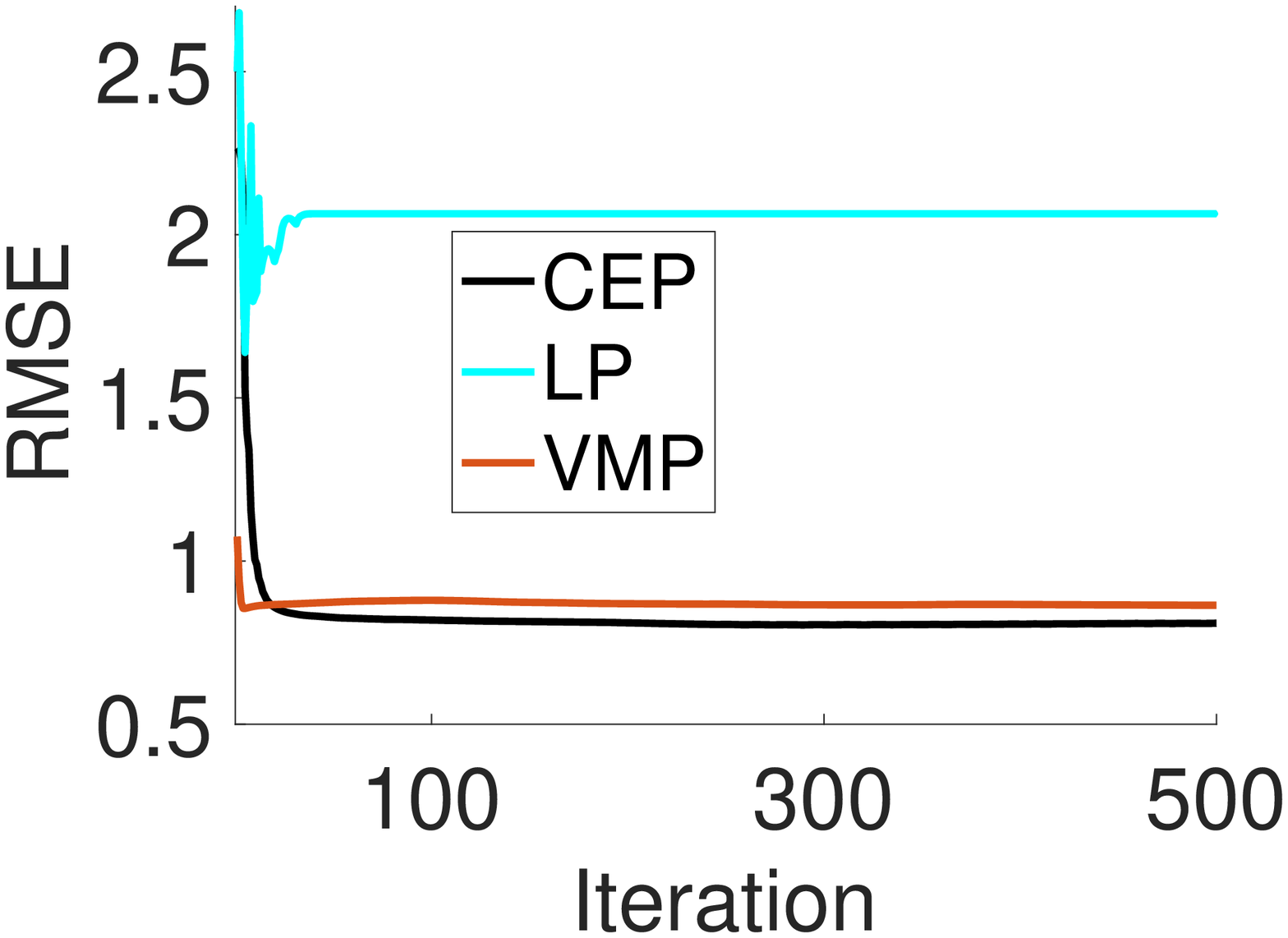}
			\caption{\textit{Alog}}
		\end{subfigure}
		&
		\begin{subfigure}[t]{0.25\textwidth}
			\centering
			\includegraphics[width=\textwidth]{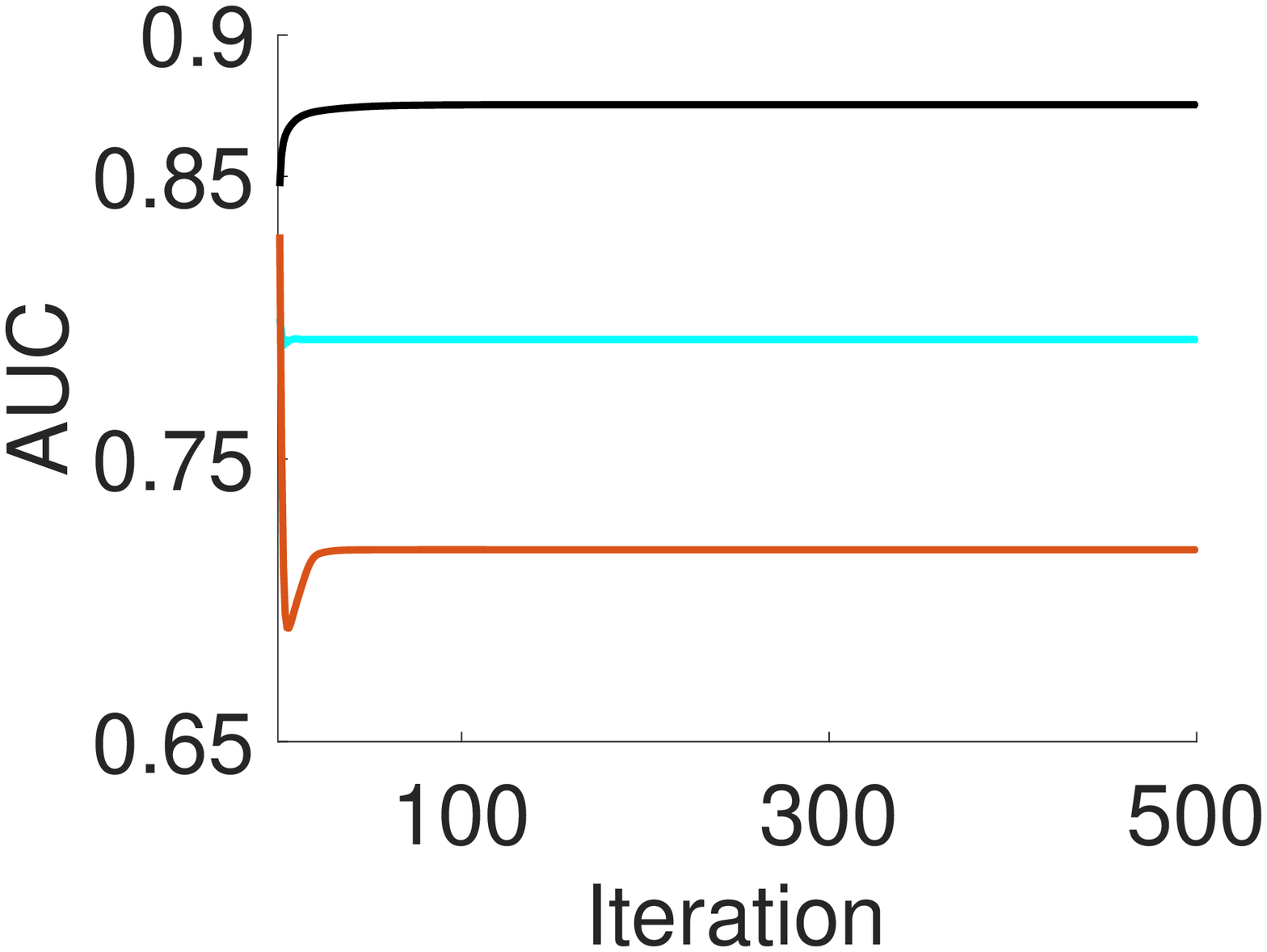}
			\caption{\textit{Enron}}
		\end{subfigure}  
	\end{tabular}
	\vspace{-0.1in}
	\caption{Average prediction accuracy \textit{v.s.} running iteration. The rank of the embeddings is $10$.}
	\label{fig:tf-iter}
	\vspace{-0.3in}
\end{figure}
\vspace{-0.1in}
\subsection{BAYESIAN TENSOR DECOMPOSITION}
\vspace{-0.1in}
We next examined our approach in Bayesian tensor decomposition (as in Section \ref{sec:btd}). Due to the production term in the likelihood (see \eqref{eq:ll_c}\eqref{eq:ll_b}),  the moment matching is intractable and hence standard EP is infeasible.  To the best of our knowledge, efficient and accurate approximations remain absent. 

\textbf{Datasets.} We used two real-world datasets~\citep{zhe2016distributed}: (1) \textit{Alog}, a $200 \times 100 \times 200$ tensor with continuous entry values, and (2) \textit{Enron}, a binary tensor of size $203 \times 203 \times 200$. \textit{Alog} represents the three-way (user, action,
resource) interactions in a file access log, and includes $0.33\%$ nonzero entries. \textit{Enron} describes the three-way relationship (sender, receiver, time) in emails, with $0.01\%$ nonzero entries. 

\textbf{Methods.} We compared with LP, variational message passing (VMP), and a state-of-the-art tensor decomposition algorithm, CP-WOPT~\citep{acar2011scalable}. VMP has been proven successful in matrix decomposition when the EP-based messages are unavailable~\citep{stern2009matchbox}. We implemented VMP for tensor decomposition. As in  ~\citep{stern2009matchbox}, to ensure tractable updates, VMP uses an augmented model representation for binary tensors, where for each entry $\bi$, a latent continuous entry value $z_\bi$ is first sampled from $\N\big(z_\bi| \1^\top (\u^1_{i_1} \circ \ldots \circ \u^K_{i_K}), 1\big),$ and then the observation $y_\bi$ determined from a step  function, $\mathds{1}(y_\bi=1)\mathds{1}(z_\bi>0) + \mathds{1}(y_\bi=0)\mathds{1}(z_\bi\le 0)$ where $\mathds{1}(\cdot)$ is the indicator function. We implemented our approach, CEP, based on the first-order Taylor approximations in \eqref{eq:1st} (the second-order based one turns out be to worse although with more computation).

The test setting is the same as in ~\citep{zhe2016distributed}. We evaluated all the methods via a $5$-fold cross validation. Specifically, we randomly split the nonzero entries into $5$ folds where $4$ folds and the same number of zero entries were sampled for training. We used the rest of nonzero entries and sampled $0.1\%$ of the remaining zero entries for test. In so doing, the nonzero and zero entries were considered as equally important, and we avoided the evaluation being dominated by the large portion of zeros. We chose the rank, \ie the dimension of the embedding vectors,  from $\{3,5,8,10\}$.   

We report the average root-mean-square-error (RMSE) for \textit{Alog} and AUC for \textit{Enron} as well as their standard deviations in Fig. \ref{fig:tf} a and c. We can see that CEP outperforms LP and CP-WOPT by a large margin in all the cases. The prediction accuracy of CEP is close to (when the rank is $3$ or $5$  for \textit{Alog}) or significantly better than VMP (in all the other cases). Note that the inferior performance of VMP in binary data might be due to the extra variational approximations for the latent continuous variables in the augmented model representation. The reason why LP is far worse than CEP might be that the numerical optimizations were saturated in poor local maxima, resulting in inferior Laplace approximations. 
We also report the per-iteration running time of CEP, LP and VMP in Fig. \ref{fig:tf} b and d. 
As we can see,  CEP has a close speed to VMP, and is much faster than LP --- on average with over $40$ times speedup. The results consistently demonstrate the advantage of CEP in both predictive performance and computational efficiency. Finally, we show in Fig. \ref{fig:tf-iter} how the predictive performance of CEP, LP and VMP varied along with the running iterations when the rank is $10$ (more results are given in the supplementary material). As we can see,  the prediction accuracy of all the methods  converges quickly and keeps stable with more iterations.
 \vspace{-0.1in}
\subsection{STREAMING TENSOR DECOMPOSITION}
\begin{figure}
	\vspace{-0.2in}
	\centering
	\setlength\tabcolsep{0.1pt}
	\begin{tabular}[c]{c}
		\begin{subfigure}[t]{0.3\textwidth}
			\centering
			\includegraphics[width=\textwidth]{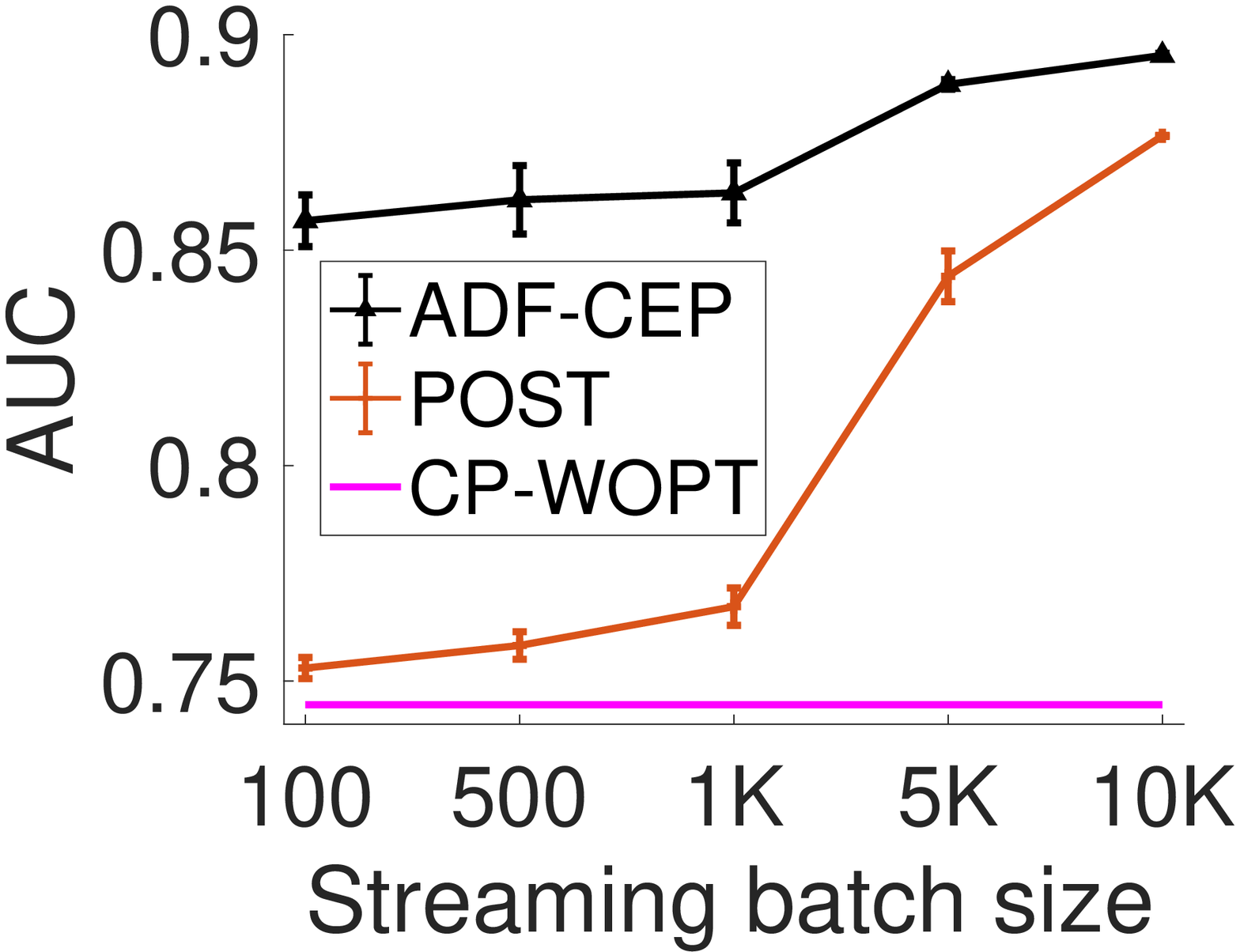}
		\end{subfigure}
	\end{tabular}
	\vspace{-0.15in}
	\caption{The predictive performance for streaming decomposition. The rank of the embeddings is $8$. \cmt{The number of latent factors is $8$.}} 
	\label{fig:stream}
	\vspace{-0.3in}
\end{figure}

\vspace{-0.1in}
Finally, we adapted CEP to the assumed density filtering (ADF) framework, denoted by ADF-CEP,  for streaming tensor decomposition. We compared with the state-of-the-art Bayesian streaming tensor decomposition algorithm, POST~\citep{du2018probabilistic}, which is essentially VMP adjusted to the streaming variational Bayes ~\citep{broderick2013streaming}. We also compared with CP-WOPT that performs static decomposition. We tested on a real-world binary tensor, \textit{DBLP}~\citep{du2018probabilistic}, which depicts the bibliography relationships (author, conference, keyword). The tensor is $10K \times 200 \times 10K$, including $\%0.001$ nonzero entries.  
We followed~\citep{du2018probabilistic} to sample $80\%$ nonzero entries and randomly sampled the same number of zero entries to obtain a balanced training set. We then sampled 50 test sets from the remaining entries --- each comprises $200$ nonzero and $1,800$ zero elements. We randomly shuffled the training entries, and partitioned them into many small batches. These batches were streamed to ADF-CEP and POST.  After all the batches were processed, we evaluated the predictive performance of the learned models on the 50 test sets. We varied the batch sizes from $\{100, 500, 1K, 5K, 10K\}$, and set the rank of the embeddings to $8$. \cmt{the number of latent factors to $8$. }We conducted  $5$ runs for each method, and report the average and standard deviation of AUC in Fig. \ref{fig:stream}. As we can see, ADF-CEP consistently outperforms POST and CP-WOPT by a large margin. 

\vspace{-0.1in}
\section{CONCLUSION}
\vspace{-0.1in}
We have developed  conditional expectation propagation (CEP) to overcome the practical barriers of EP when the moment matching is intractable. CEP conducts efficient and analytical updates, without the need for hand-crafted approximations,  expensive sampling or numerical optimizations. The performance of CEP in the inference tasks of several important Bayesian models is encouraging. In the future work, we will continue to explore CEP in theory, \eg energy functions, and apply CEP in more complex models, \eg Bayesian neural networks.

\bibliographystyle{apalike}
\bibliography{CEP}

\begin{thebibliography}{}

\bibitem[Acar et~al., 2011]{acar2011scalable}
Acar, E., Dunlavy, D.~M., Kolda, T.~G., and M{\o}rup, M. (2011).
\newblock Scalable tensor factorizations for incomplete data.
\newblock {\em Chemometrics and Intelligent Laboratory Systems}, 106(1):41--56.

\bibitem[Boyen and Koller, 1998]{boyen1998tractable}
Boyen, X. and Koller, D. (1998).
\newblock Tractable inference for complex stochastic processes.
\newblock In {\em Proceedings of the Fourteenth conference on Uncertainty in
  artificial intelligence}, pages 33--42. Morgan Kaufmann Publishers Inc.

\bibitem[Broderick et~al., 2013]{broderick2013streaming}
Broderick, T., Boyd, N., Wibisono, A., Wilson, A.~C., and Jordan, M.~I. (2013).
\newblock Streaming variational bayes.
\newblock In {\em Advances in Neural Information Processing Systems}, pages
  1727--1735.

\bibitem[Du et~al., 2018]{du2018probabilistic}
Du, Y., Zheng, Y., Lee, K.-c., and Zhe, S. (2018).
\newblock Probabilistic streaming tensor decomposition.
\newblock In {\em 2018 IEEE International Conference on Data Mining (ICDM)},
  pages 99--108. IEEE.

\bibitem[Du\v{s}ek, 2013]{probitEP}
Du\v{s}ek, O. (2013).
\newblock Expectation propagation for a probit regression model.
\newblock Technical report, Institute of Formal and Applied Linguistics.

\bibitem[Eskin et~al., 2004]{eskin2004laplace}
Eskin, E., Smola, A.~J., and Vishwanathan, S. (2004).
\newblock Laplace propagation.
\newblock In {\em Advances in neural information processing systems}, pages
  441--448.

\bibitem[Eslami et~al., 2014]{eslami2014just}
Eslami, S.~A., Tarlow, D., Kohli, P., and Winn, J. (2014).
\newblock Just-in-time learning for fast and flexible inference.
\newblock In {\em Advances in Neural Information Processing Systems}, pages
  154--162.

\bibitem[Gelman et~al., 2013]{gelman2013bayesian}
Gelman, A., Stern, H.~S., Carlin, J.~B., Dunson, D.~B., Vehtari, A., and Rubin,
  D.~B. (2013).
\newblock Bayesian data analysis.
\newblock chapter~10. Chapman and Hall/CRC.

\bibitem[Graepel et~al., 2010]{graepel2010web}
Graepel, T., Candela, J.~Q., Borchert, T., and Herbrich, R. (2010).
\newblock Web-scale {B}ayesian click-through rate prediction for sponsored
  search advertising in {M}icrosoft's {B}ing search engine.
\newblock Omnipress.

\bibitem[Harshman, 1970]{Harshman70parafac}
Harshman, R.~A. (1970).
\newblock Foundations of the {PARAFAC} procedure: Model and conditions for
  an''explanatory''multi-mode factor analysis.
\newblock {\em UCLA Working Papers in Phonetics}, 16:1--84.

\bibitem[Heess et~al., 2013]{heess2013learning}
Heess, N., Tarlow, D., and Winn, J. (2013).
\newblock Learning to pass expectation propagation messages.
\newblock In {\em Advances in Neural Information Processing Systems}, pages
  3219--3227.

\bibitem[Herbrich et~al., 2007]{NIPS2006_3079}
Herbrich, R., Minka, T., and Graepel, T. (2007).
\newblock Trueskill\texttrademark : A bayesian skill rating system.
\newblock In Sch\"{o}lkopf, B., Platt, J.~C., and Hoffman, T., editors, {\em
  Advances in Neural Information Processing Systems 19}, pages 569--576. MIT
  Press.

\bibitem[Hern{\'a}ndez-Lobato and Adams, 2015]{hernandez2015probabilistic}
Hern{\'a}ndez-Lobato, J.~M. and Adams, R. (2015).
\newblock Probabilistic backpropagation for scalable learning of bayesian
  neural networks.
\newblock In {\em International Conference on Machine Learning}, pages
  1861--1869.

\bibitem[Hoffman and Gelman, 2014]{hoffman2014no}
Hoffman, M.~D. and Gelman, A. (2014).
\newblock The no-u-turn sampler: adaptively setting path lengths in hamiltonian
  monte carlo.
\newblock {\em Journal of Machine Learning Research}, 15(1):1593--1623.

\bibitem[Jitkrittum et~al., 2015]{jitkrittum2015kernel}
Jitkrittum, W., Gretton, A., Heess, N., Eslami, S., Lakshminarayanan, B.,
  Sejdinovic, D., and Szab{\'o}, Z. (2015).
\newblock Kernel-based just-in-time learning for passing expectation
  propagation messages.
\newblock In {\em Proceedings of the Thirty-First Conference on Uncertainty in
  Artificial Intelligence}, pages 405--414. AUAI Press.

\bibitem[Kschischang et~al., 2001]{kschischang2001factor}
Kschischang, F.~R., Frey, B.~J., and Loeliger, H.-A. (2001).
\newblock Factor graphs and the sum-product algorithm.
\newblock {\em IEEE Transactions on information theory}, 47(2):498--519.

\bibitem[Lauritzen, 1992]{lauritzen1992propagation}
Lauritzen, S.~L. (1992).
\newblock Propagation of probabilities, means, and variances in mixed graphical
  association models.
\newblock {\em Journal of the American Statistical Association},
  87(420):1098--1108.

\bibitem[Li et~al., 2015]{li2015stochastic}
Li, Y., Hern{\'a}ndez-Lobato, J.~M., and Turner, R.~E. (2015).
\newblock Stochastic expectation propagation.
\newblock In {\em Advances in Neural Information Processing Systems}, pages
  2323--2331.

\bibitem[Maybeck, 1982]{maybeck1982stochastic}
Maybeck, P.~S. (1982).
\newblock {\em Stochastic models, estimation, and control}, volume~3.
\newblock Academic press.

\bibitem[Minka, 2004]{minka2004power}
Minka, T. (2004).
\newblock Power {EP}.
\newblock Technical report, Technical report, Microsoft Research, Cambridge.

\bibitem[Minka et~al., 2014]{minka2014infer}
Minka, T., Winn, J., Guiver, J., and Knowles, D. (2014).
\newblock Infer .net 2.4, 2010. microsoft research cambridge.

\bibitem[Minka, 2001]{minka2001expectation}
Minka, T.~P. (2001).
\newblock Expectation propagation for approximate bayesian inference.
\newblock In {\em Proceedings of the Seventeenth conference on Uncertainty in
  artificial intelligence}, pages 362--369. Morgan Kaufmann Publishers Inc.

\bibitem[Murphy et~al., 1999]{murphy1999loopy}
Murphy, K.~P., Weiss, Y., and Jordan, M.~I. (1999).
\newblock Loopy belief propagation for approximate inference: An empirical
  study.
\newblock In {\em Proceedings of the Fifteenth conference on Uncertainty in
  artificial intelligence}, pages 467--475. Morgan Kaufmann Publishers Inc.

\bibitem[Stern et~al., 2009]{stern2009matchbox}
Stern, D.~H., Herbrich, R., and Graepel, T. (2009).
\newblock Matchbox: large scale online {B}ayesian recommendations.
\newblock In {\em Proceedings of the 18th international conference on World
  wide web}, pages 111--120. ACM.

\bibitem[Zhe et~al., 2016a]{zhe2016online}
Zhe, S., Lee, K.-c., Zhang, K., and Neville, J. (2016a).
\newblock Online spike-and-slab inference with stochastic expectation
  propagation.
\newblock {\em NIPS WS}.

\bibitem[Zhe et~al., 2016b]{zhe2016distributed}
Zhe, S., Zhang, K., Wang, P., Lee, K.-c., Xu, Z., Qi, Y., and Ghahramani, Z.
  (2016b).
\newblock Distributed flexible nonlinear tensor factorization.
\newblock In {\em Advances in Neural Information Processing Systems}, pages
  928--936.

\end{thebibliography}

\section*{Supplementary Materials}
In this extra material, we show the details of the updates in our doubly stochastic variational EM algorithm in Section 1, and the clusters of latent factors discovered by our model in Section 2. 
\section{CEP for Bayesian Tensor Decomposition, Logistic Regression and Probit Regression}

\subsection{Bayesian Tensor Decomposition}

In this section, we present the details of posterior inference for Bayesian tensor decomposition based on conditional EP, with observed data entries $S$. We use the same notations as those in section 4.1.

\subsubsection{Continuous Tensor}

Let us first consider continuous entry values $\{y_\bi\}_{\bi \in S}$. The joint probability of Bayesian tensor decomposition model, according to (12) in section 4.1, is given by
\begin{equation*}
    \begin{split}
        p(\{y_\bi\}_{\bi \in S}, \Ucal, \tau) = & \; \mathrm{Gam}(\tau|a_0, b_0) \prod_{k = 1}^K \prod_{s = 1}^{d_k} \N(\u_s^k| \m_s^k, v\I)\\
                                                \cdot & \; \prod_{\bi \in S} \N(y_\bi |  \1^\top (\u^1_{i_1} \circ \ldots \circ \u^K_{i_K}), \tau^{-1}).
    \end{split}
\end{equation*}
And the posterior can be easily derived as
\begin{equation*}
    \begin{split}
        p(\Ucal, \tau |\{y_\bi\}_{\bi \in S})  \propto & \; \mathrm{Gam}(\tau|a_0, b_0) \prod_{k = 1}^K \prod_{s = 1}^{d_k} \N(\u_s^k| \m_s^k, v\I)\\
                                                \cdot & \; \prod_{\bi \in S} \N(y_\bi |  \1^\top (\u^1_{i_1} \circ \ldots \circ \u^K_{i_K}), \tau^{-1}).
    \end{split}
\end{equation*}
Since priors of the embeddings $\Ucal$ and $\tau$ are already in expenontial family, we only need to estimate the likelihoods. By using fully factorized approximation
\begin{equation*}
    \N(y_\bi |  \1^\top (\u^1_{i_1} \circ \ldots \circ \u^K_{i_K}), \tau^{-1}) \approx \tf_\bi(\tau)\prod_{k=1}^K  \tf_{\bi}^k(\u_{i_k}^k),
\end{equation*}
we have the approximate posterior distribution
\begin{align}
       & q(\Ucal, \tau)  \propto \mathrm{Gam}(\tau|a_0, b_0) \prod_{k = 1}^K \prod_{s = 1}^{d_k} \N(\u_s^k| \m_s^k, v\I) \nonumber \\
        & \qquad \qquad \cdot  \prod_{\bi \in S} \tf_\bi(\tau)\prod_{k=1}^K \tf_{\bi}^k(\u_{i_k}^k), \nonumber
\end{align}
where $\tf_\bi(\tau) = \mathrm{Gam}(\tau | a_\bi, b_\bi)$ and $\tf_{\bi}^k(\u_{i_k}^k) = \N(\u_{i_k}^k | \m_\bi^k, \S_\bi^k)$. 

\noindent Accordingly, the cavity distribution is
\begin{equation*}
        q^{\setminus \bi}(\Ucal, \tau) \propto  \frac{q(\Ucal, \tau) }{\tf_\bi(\tau)\prod_{k=1}^K \tf_{\bi}^k(\u_{i_k}^k)},
\end{equation*}
and the tilted distribution is 
\begin{equation*}
    \hat{p}_\bi(\Ucal, \tau) \propto  q^{\setminus \bi}(\Ucal, \tau) \N(y_\bi |  \1^\top (\u^1_{i_1} \circ \ldots \circ \u^K_{i_K}), \tau^{-1}).
\end{equation*}

\noindent To update messages $\tf_{\bi}^k(\u_{i_k}^k)$, the conditional tilted distribution used to match moments is 
\begin{align}
    \hat{p}_\bi(\u_{i_k}^k|\u_\bi^{\setminus k}, \tau) \propto \N(\u^k_{i_k}|\m^k_{i_k}, \S^k_{i_k}) \N(y_\bi | {\z_\bi^{\setminus k}}^\top \u^k_{i_k}, \tau^{-1}), \nonumber
\end{align}
where
\begin{align}
    & \S_{i_k}^k =  \left( \sum_{\bj \in S,\; \bj \neq \bi,\; j_k = i_k} {\S_{\bj}^k}^{-1} + v\I\right)^{-1}, \nonumber \\
    & \m_{i_k}^k = \S_{i_k}^k \left( \sum_{\bj \in S,\; \bj \neq \bi,\; j_k = i_k} {\S_{\bj}^k}^{-1} \m_\bj^k  + v\m_{s=i_k}^k\right) .\nonumber
\end{align}
To clearify the notation, $\m_s^k$ are means in prior probability. 

\noindent Thus, the conditional moments are
\begin{align}
    &\mathrm{cov}(\u_{i_k}^k|\u_\bi^{\setminus k}, \tau) = \big[{\S^k_{i_k}}^{-1} + \tau (\z_\bi^{\setminus k} {\z_\bi^{\setminus k}}^\top)\big]^{-1}, \nonumber \\  
    &\EE(\u_{i_k}^k|\u_\bi^{\setminus k}, \tau ) = \mathrm{cov}(\u_{i_k}^k|\u_\bi^{\setminus k}, \tau)\big[ {\S^k_{i_k}}^{-1}\m^k_{i_k} + \tau y_\bi \z_\bi^{\setminus k}\big] \nonumber.
\end{align}

\noindent Because of fully-factorized approximate distribution, we have
\begin{align}
    &\EE_{q}(\tau) = \frac{a_0 + \sum_{\bi \in S} a_\bi - |S|}{b_0 + \sum_{\bi \in S} b_\bi}, \nonumber \\
    &\EE_{q}(\z_\bi^{\setminus k}) = \EE_{q}(\u_{i_1}^1) \circ  \ldots \circ \EE_{q}(\u_{i_{k-1}}^{k-1}) \nonumber \\
    & \qquad \qquad \circ \EE_{q}(\u_{i_{k+1}}^{k+1})\circ \ldots \circ \EE_{q}(\u_{i_K}^K), \nonumber \\
    &\EE_{q}(\z_\bi^{\setminus k}{\z_\bi^{\setminus k}}^\top) = \EE_{q}(\u_{i_1}\u_{i_1}^\top) \circ  \ldots \circ \EE_{q}(\u_{i_{k-1}}^{k-1}{\u_{i_{k-1}}^{k-1}}^\top) \nonumber\\
    &\qquad \qquad \qquad \circ \EE_{q}(\u_{i_{k+1}}^{k+1}{\u_{i_{k+1}}^{k+1}}^\top)\circ \ldots \circ \EE_{q}(\u_{i_K}^K{\u_{i_K}^K}^\top), \nonumber
\end{align}
where
\begin{align}
    &\mathrm{cov}_{q}(\u_{i_k}^k) = \left( \sum_{\bj \in S,\; j_k = i_k} {\S_{\bj}^k}^{-1} + v\I\right)^{-1}, \nonumber \\
    &\EE_{q}(\u_{i_k}^k) = \mathrm{cov}_{q}(\u_{i_k}^k) \left( \sum_{\bj \in S,\; j_k = i_k} {\S_{\bj}^k}^{-1} \m_\bj^k  + v\m_{s=i_k}^k\right) ,\nonumber \\
    &\EE_{q}(\u_{i_k}\u_{i_k}^\top) = \mathrm{cov}_{q}(\u_{i_k}^k) + \EE_{q}(\u_{i_k}^k){\EE_{q}(\u_{i_k}^k)}^\top \nonumber.
\end{align}

\noindent By taking the expectation over the first-order Taylor approximation w.r.t $q(\u_\bi^{\setminus k}, \tau)$, we can just replace $\tau$, $\z_\bi^{\setminus k}$ and $\z_\bi^{\setminus k} {\z_\bi^{\setminus k}}^\top$ with $\EE(\tau)$, $\EE(\z_\bi^{\setminus k})$ and $\EE(\z_\bi^{\setminus k} {\z_\bi^{\setminus k}}^\top)$ to match moments and 
update messages $\tf_{\bi}^k(\u_{i_k}^k)$. And the updated $\m_\bi^k$ and  $\S_\bi^k$ are 
\begin{align}
    &{\S_\bi^k}^* = \left(\EE_{q}(\tau)\EE_{q}(\z_\bi^{\setminus k}{\z_\bi^{\setminus k}}^\top)\right)^{-1} ,\nonumber \\
    &{\m_\bi^k}^* = {\S_\bi^k}^*\left(y_\bi\EE_{q}(\tau)\EE_{q}(\z_\bi^{\setminus k})\right)    .\nonumber
\end{align}

\noindent As for updating $\tf_\bi(\tau)$, the conditional tilted distribution for $\tau$ is 
\begin{align}
    & \hat{p}_\bi(\tau|\u_{i_1}^1, \ldots, \u_{i_K}^K) \nonumber \\
    \propto & \; \mathrm{Gam}(\tau|a^{\setminus \bi}, b^{\setminus \bi}) \N(y_\bi |  \1^\top (\u^1_{i_1} \circ \ldots \circ \u^K_{i_K}), \tau^{-1}) \nonumber \\
    = &\; \mathrm{Gam}(\tau|\hat{a}, \hat{b}) \nonumber,
\end{align}
where
\begin{align}
    & a^{\setminus \bi} = a_0 + \sum_{\bj \in S, \bj \neq \bi} a_\bj - |S| + 1,\nonumber \\
    & b^{\setminus \bi} = b_0 + \sum_{\bj \in S, \bj \neq \bi} b_\bj,\nonumber \\
    & \hat{a} = a^{\setminus \bi} + \frac{1}{2},\nonumber \\
    & \hat{b}= b^{\setminus \bi} + \frac{1}{2} (y_\bi - \1^\top (\u^1_{i_1} \circ \ldots \circ \u^K_{i_K}))^2.\nonumber
\end{align}
We only need to take expectation over the first-order Taylor approximation of $\hat{b}$ w.r.t $q(\u_\bi^{\setminus k}, \tau)$ so as to update messages $\tf_\bi(\tau)$. And the 
updated $a_\bi$ and $b_\bi$ are
\begin{align}
    & a_\bi^* = \frac{3}{2}, \nonumber \\
    & b_\bi^* = \frac{1}{2}(y_\bi - \1^\top [\EE(\u^1_{i_1}) \circ \ldots \circ \EE(\u^K_{i_K})])^2. \nonumber
\end{align}
\subsubsection{Binary Tensor}

When entry values are bianry values, accroding to (13) in section 4.1, the joint probability becomes
\begin{equation*}
    \begin{split}
        p(\{y_\bi\}_{\bi \in S}, \Ucal) = & \;  \prod_{k = 1}^K \prod_{s = 1}^{d_k} \N(\u_s^k| \m_s^k, v\I)\\
                                                \cdot & \; \prod_{\bi \in S} \psi\big((2y_\bi -1)\1^\top (\u^1_{i_1} \circ \ldots \circ \u^K_{i_K})\big) .
    \end{split}
\end{equation*}
And the corresponding posterior distribution and approximate posterior distribution using fully-factorized approximation are
\begin{align}
    & p(\Ucal |\{y_\bi\}_{\bi \in S})  \propto \prod_{k = 1}^K \prod_{s = 1}^{d_k} \N(\u_s^k| \m_s^k, v\I) \nonumber\\
    & \qquad \qquad \qquad \cdot \prod_{\bi \in S} \psi\big((2y_\bi -1)\1^\top (\u^1_{i_1} \circ \ldots \circ \u^K_{i_K})\big), \nonumber \\
    & q(\Ucal) = \prod_{k = 1}^K \prod_{s = 1}^{d_k} \N(\u_s^k| \m_s^k, v\I) \prod_{\bi \in S} \prod_{k=1}^K \tf_{\bi}^k(\u_{i_k}^k), \nonumber
\end{align}
where $\psi(\cdot)$ is the cumulative density function (CDF) of the standard normal distribution and $\tf_{\bi}^k(\u_{i_k}^k) = \N(\u_{i_k}^k | \m_\bi^k, \S_\bi^k)$.

\noindent Cavity distribution and tilted distribution can also be derived as
\begin{align}
    &q^{\setminus \bi}(\Ucal) \propto   \frac{q(\Ucal, \tau) }{\prod_{k=1}^K \tf_{\bi}^k(\u_{i_k}^k)}, \nonumber \\
    &\hat{p}_\bi(\Ucal) \propto   q^{\setminus \bi}(\Ucal) \psi\big((2y_\bi -1)\1^\top (\u^1_{i_1} \circ \ldots \circ \u^K_{i_K})\big). \nonumber
\end{align}

\noindent To update messages $\tf_{\bi}^k(\u_{i_k}^k)$, the conditional tilted distribution for $\u_{i_k}^k$ is
\begin{align}
    & \; \hat{p}_\bi(\u_{i_k}^k|\u_\bi^{\setminus k}, \tau) \nonumber \\
    \propto & \; \N(\u^k_{i_k}|\m^k_{i_k}, \S^k_{i_k}) \psi\big((2y_\bi -1)\1^\top (\u^1_{i_1} \circ \ldots \circ \u^K_{i_K})\big), \nonumber
\end{align}
where
\begin{align}
    & \S_{i_k}^k =  \left( \sum_{\bj \in S,\; \bj \neq \bi,\; j_k = i_k} {\S_{\bj}^k}^{-1} + v\I\right)^{-1}, \nonumber \\
    & \m_{i_k}^k = \S_{i_k}^k \left( \sum_{\bj \in S,\; \bj \neq \bi,\; j_k = i_k} {\S_{\bj}^k}^{-1} \m_\bj^k  + v\m_{s=i_k}^k\right) .\nonumber
\end{align}
\noindent It is tricky to calculate the conditional moments of $\hat{p}_\bi(\u_{i_k}^k|\u_\bi^{\setminus k})$. 

\noindent Let 
\begin{align}
    & \int \N(\u^k_{i_k}|\m^k_{i_k}, \S^k_{i_k}) \psi\big((2y_\bi -1)\1^\top (\u^1_{i_1} \circ \ldots \circ \u^K_{i_K})\big) d\u^k_{i_k} \nonumber \\
    = & \psi \left( \frac{(2y_\bi - 1){\z_\bi^{\setminus k}}^\top \m^k_{i_k}}{\sqrt{1 + {\z_\bi^{\setminus k}}^\top \S^k_{i_k}  \z_\bi^{\setminus k}}} \right) \nonumber\\
    = & Z \nonumber 
\end{align}
Then, the conditional moments are
\begin{align}
    &\mathrm{cov}(\u_{i_k}^k|\u_\bi^{\setminus k}) = \S^k_{i_k} -  \S^k_{i_k} \A \S^k_{i_k} , \nonumber \\  
    &\EE(\u_{i_k}^k|\u_\bi^{\setminus k}) = \m_{i_k}^k + \S^k_{i_k}\frac{\partial \log Z}{\partial \m_{i_k}^k}, \nonumber
\end{align}
with
\begin{align}
    \A = \frac{\partial \log Z}{\partial \m_{i_k}^k} (\frac{\partial \log Z}{\partial \m_{i_k}^k})^\top -  2\frac{\partial \log Z}{\partial \S_{i_k}^k}. \nonumber
\end{align}
Finally, by taking the expectation of the first-order Taylor approximation w.r.t $q(\u_\bi^{\setminus k})$ and replacing $\z_\bi^{\setminus k}$ and ${\z_\bi^{\setminus k}}^\top \S^k_{i_k}  \z_\bi^{\setminus k}$ with $\EE(\z_\bi^{\setminus k})$ and $tr\left(\S^k_{i_k}\EE(\z_\bi^{\setminus k}{\z_\bi^{\setminus k}}^\top)\right)$, the updated $\m_\bi^k$ and $\S_\bi^k$ are
\begin{align} 
    & {\S_\bi^k}^* = \left( T_1- {\S^k_{i_k}}^{-1} \right)^{-1}, \nonumber \\
    & {\m_\bi^k}^* = {\S_\bi^k}^*\left( T_2 -  {\S^k_{i_k}}^{-1}\m^k_{i_k} \right), \nonumber
\end{align}
with
\begin{align}
    & T_1 =  \EE_q(\mathrm{cov}(\u_{i_k}^k|\u_\bi^{\setminus k}))^{-1}, \nonumber \\
    & T_2 = \EE_{q}(\mathrm{cov}(\u_{i_k}^k|\u_\bi^{\setminus k}))^{-1} \EE_{q}(\EE(\u_{i_k}^k|\u_\bi^{\setminus k})). \nonumber
\end{align}

\subsection{Logistic Regression}
In this section, we show the details of message updating based on fully-factorized approximation for logistic regression in standard EP and conditional EP respectively. We use the same notations as those in section 4.2.

\noindent As described in section 4.2, the joint probability for Bayesian logistic regression is
\begin{align}
    p(\y, \w|\X) = p(\w) \prod_{i=1}^n 1/\big(1  + \exp(-(2y_i - 1)\w^\top \x_i)\big),  \nonumber 
\end{align}
where $p(\w) = \N(\w| \0, \lambda \I)$.

\noindent The corresponding approximate posterior distribution based on fully-factorized approximation is
\begin{align}
    q(\w) \propto p(\w) \prod_{i=1}^n \prod_m\tf_{im}(w_m) \nonumber
\end{align}
where $\tf_{im}(w_m) = \N(w_m|\mu_{im}, v_{im})$.

\noindent The remaining two related distributions, cavity distribution and tilted distribution, can also be derived as 
\begin{align}
    & q^{\setminus i}(\w) \propto p(\w) \prod_{j \neq i} \prod_m\tf_{jm}(w_m) = \prod_m \N(w_m|\mu_m^{\setminus i}, v_m^{\setminus i}), \nonumber \\
    & \hat{p}_i(\w) \propto \frac{1}{1  + \exp(-(2y_i - 1)\w^\top \x_i)} q^{\setminus i}(\w), \nonumber
\end{align}
where
\begin{align}
    &v_m^{\setminus i} = \frac{1}{1/\lambda + \sum_{j \neq i} 1/v_{jm}}, \nonumber \\
    &\mu_m^{\setminus i} = v_m^{\setminus i}\sum_{j \neq i} \frac{\mu_{jm}}{v_{jm}}. \nonumber
\end{align}
\subsubsection{Standard EP for Logistic Regression}
To update the  message $\tf_{im}(w_m)$ , we first project the cavity distribution $q^{\setminus i}(\w)$ into the two-dimensional subspace represented by the data vector $\x_i$. The projected distribution is a two-dimensional normal with mean
and variance,
\begin{align}
    & M_m^{\setminus i} = \Big(\begin{matrix}
        x_{im} \mu_m^{\setminus i} \\
        \sum_{l \neq m} x_{il} \mu_l^{\setminus i} 
    \end{matrix} \Big), \nonumber\\
    & V_m^{\setminus i} = \Big( \begin{matrix}
        x_{im}^2 v_m^{\setminus i} & 0 \\
        0 & \sum_{l \neq m} x_{il}^2 v_l^{\setminus i}
     \end{matrix}\Big).\nonumber 
\end{align}

\noindent Define the (unnormalized) tilted distribution of $\boldeta = \big( \begin{matrix}
    \eta_1 \\
    \eta_2
\end{matrix}\big)$,
\begin{align}
    &\frac{1}{1  + \exp(-(2y_i - 1)(\eta_1 + \eta_2))}q^{\setminus i}(\boldeta) \nonumber \\
    =& \frac{1}{1  + \exp(-(2y_i - 1)(\eta_1 + \eta_2))}\N(\boldeta|M_m^{\setminus i}, V_m^{\setminus i}). \nonumber
\end{align}

\noindent  To compute the moments 0, 1, 2 of this unnormalized tilted distribution of $\boldeta$,
\begin{align}
   & E_0 = \int \frac{\N(\boldeta|M_m^{\setminus i}, V_m^{\setminus i})}{1  + \exp(-(2y_i - 1)(\eta_1 + \eta_2))} d \boldeta , \nonumber \\
   & E_1 = \int  \boldeta \frac{\N(\boldeta|M_m^{\setminus i}, V_m^{\setminus i})}{1  + \exp(-(2y_i - 1)(\eta_1 + \eta_2))} d \boldeta , \nonumber \\
   & E_2 = \int \boldeta \boldeta^\top \frac{\N(\boldeta|M_m^{\setminus i}, V_m^{\setminus i})}{1  + \exp(-(2y_i - 1)(\eta_1 + \eta_2))} d \boldeta,\nonumber
\end{align}
we use numerical integral method, Gauss-Hermite quadrature, with given nodes and weights $\{(\gamma_j, \alpha_j)\}$,
\begin{align}
    & E_0 \approx \sum_i \sum_j  \frac{\alpha_i \alpha_j}{1  + \exp(-(2y_i - 1)(\gamma_i + \gamma_j))}, \nonumber \\
    & E_1 \approx \sum_i \sum_j  \frac{\alpha_i  \alpha_j \big( \begin{matrix}
        \gamma_i \\
        \gamma_j
    \end{matrix}\big)}{1  + \exp(-(2y_i - 1)(\gamma_i + \gamma_j))}, \nonumber \\
    &E_2 \approx \sum_i \sum_j  \frac{\alpha_i  \alpha_j \big( \begin{matrix}
        \gamma_i \\
        \gamma_j
    \end{matrix}\big) \big( \begin{matrix}
        \gamma_i \\
        \gamma_j
    \end{matrix}\big)^\top}{1  + \exp(-(2y_i - 1)(\gamma_i + \gamma_j))}. \nonumber 
\end{align}
After computing the mean and variance of the tilted distribution,
\begin{align}
    & M_m = \frac{E_1}{E_0}, \nonumber \\
    & V_m = \frac{E_2}{E_0} - \left(\frac{E_1}{E_0}\right)\left(\frac{E_1}{E_0}\right)^\top \nonumber,  
\end{align}
subtract off the cavity distribution to get the moments of the updated approximating factor $\N(\boldeta|M_m^{i}, V_m^{i})$,
\begin{align}
    & {V_m^{i}}^{-1}M_m^{i}= {V_m}^{-1}M_m - {V_m^{\setminus i}}^{-1} M_m^{\setminus i}, \nonumber \\
    & {V_m^{i}}^{-1} = {V_m}^{-1} - {V_m^{\setminus i}}^{-1}. \nonumber
\end{align}

\noindent To get the moments of updated factor in original space, we need to do an inverted projection and the updated $\mu_{im}$ and $v_{im}$ are
\begin{align}
    & v_{im}^* = \frac{(V_m^{i})_{1,1}}{x_{im}^2}, \nonumber \\
    & \mu_{im}^* = v_{im}^*x_{im}({V_m^{i}}^{-1}M_m^{i})_{1,1}. \nonumber
\end{align}

\subsubsection{Conditional EP for Logistic Regression}
To update $\tf_{im}(w_m)$, the conditional tilted distribution is 
\begin{align}
    \hat{p}_i(w_m|\w_{\setminus m}) \propto q^{\setminus i}(w_m) g_{im}(w_m|\w_{\setminus m}), \nonumber 
\end{align}
where
\begin{align}
    & q^{\setminus i}(w_m) = \N(w_m|\mu_m^{\setminus i}, v_m^{\setminus i}), \nonumber \\
    & g_{im}(w_m|\w_{\setminus m}) = \mathrm{logit}^{-1}((2y_i - 1)(w_mx_{im} + \w_{\setminus m}^\top {\x_i}_{\setminus m})). \nonumber
\end{align}

\noindent By using Gauss-Hermite quadrature, we can approximate the moments 0,1,2 of the unnormalized conditional tilted distribution with given quadrature nodes and weights $\{\gamma_j, \alpha_j\}$,
\begin{align}
    & E_0 = \int q^{\setminus i}(w_m) g_{im}(w_m|\w_{\setminus m}) d w_m \nonumber \\
    & \quad \approx  \sum_j \alpha_j g_{im}(\gamma_j|\w_{\setminus m}) \nonumber, \\
    &  E_1 =  \int w_m q^{\setminus i}(w_m) g_{im}(w_m|\w_{\setminus m}) d w_m \nonumber \\
    & \quad \approx\sum_j \alpha_j \gamma_j g_{im}(\gamma_j|\w_{\setminus m}) \nonumber \\
    & E_2 =  \int w_m^2 q^{\setminus i}(w_m) g_{im}(w_m|\w_{\setminus m}) d w_m \nonumber \\
    & \quad \approx \sum_j \alpha_j \gamma_j^2 g_{im}(\gamma_j|\w_{\setminus m}) \nonumber.
\end{align}

With these moments, we can easily get the conditional mean and variance,
\begin{align}
    & \EE(w_m|\w_{\setminus m}) = \frac{E_1}{E_0}, \nonumber \\
    & \mathrm{var}(w_m|\w_{\setminus m}) = \frac{E_2}{E_0} - \left(\frac{E_1}{E_0}\right)^2, \nonumber
\end{align}
and corresponding Hessian w.r.t $\w_{\setminus m}$,
\begin{align}
    & \nabla \nabla \EE(w_m|\w_{\setminus m}) = \frac{E_1 \sum_j t_j c_j - E_0\sum_j t_j \gamma_j c_j}{E_0^3} {\x_i}_{\setminus m}{\x_i}_{\setminus m}^\top \nonumber, \\
    & \nabla \nabla \mathrm{var}(w_m|\w_{\setminus m}) = \nabla \nabla \EE(w_m^2|\w_{\setminus m}) \nonumber \\
    & \qquad \qquad \qquad - 2\nabla \EE(w_m|\w_{\setminus m}) {\nabla \EE(w_m|\w_{\setminus m})}^\top \nonumber \\
    & \qquad \qquad \qquad - 2 \EE(w_m|\w_{\setminus m}) \nabla \nabla \EE(w_m|\w_{\setminus m}), \nonumber 
\end{align}
where
\begin{align}
    & \nabla \EE(w_m|\w_{\setminus m}) = \frac{E_1 \sum_j t_j - E_0 \sum_j t_j \gamma_j }{E_0^2} (2y_i - 1) {\x_i}_{\setminus m}, \nonumber \\
    & \nabla \nabla \EE(w_m^2|\w_{\setminus m}) = \frac{E_2 \sum_j t_j c_j - E_0\sum_j t_j \gamma_j^2 c_j}{E_0^3} {\x_i}_{\setminus m}{\x_i}_{\setminus m}^\top, \nonumber \\
    & t_j = \alpha_j g_{im}^2(\gamma_j|\w_{\setminus m}), \nonumber \\
    & c_j = E_0(1-2g_{im}(\gamma_j|\w_{\setminus m})) + 2\sum_j \alpha_j g_{im}^2(\gamma_j|\w_{\setminus m}). \nonumber
\end{align}

\noindent Finally, by taking the expectation over the first-order or second-order Taylor approximation w.r.t $q(\w_{\setminus m})$, we can simply replace $\w_{\setminus m}$ with $\EE_{q}(\w_{\setminus m})$ to compute the mean and variance of tilted distribution and update messages.
The updated variance and mean of $\tf_{im}(w_m)$ can be described as 
\begin{align}
    & v_{im}^* =  \left(\frac{1}{v_m} - \frac{1}{v_m^{\setminus i}}\right)^{-1}, \nonumber \\
    & \mu_{im}^* = v_{im}^*\left(\frac{\mu_m}{v_m} - \frac{\mu_m^{\setminus i}}{v_m^{\setminus i}} \right). \nonumber
\end{align}
\noindent Let 
\begin{align}
    & h_1(\w_{\setminus m}) = \EE(w_m|\w_{\setminus m}), \nonumber \\
    & h_2(\w_{\setminus m}) = \mathrm{var}(w_m|\w_{\setminus m}), \nonumber \\
    & h_3(\w_{\setminus m}) = \nabla \nabla \EE(w_m|\w_{\setminus m}), \nonumber \\
    & h_4(\w_{\setminus m}) = \nabla \nabla \mathrm{var}(w_m|\w_{\setminus m}). \nonumber
\end{align}

\noindent For fisrt-order approximation, 
\begin{align}
    & \mu_m = h_1(\EE_{q}(\w_{\setminus m})), \nonumber \\
    & v_m = h_2(\EE_{q}(\w_{\setminus m})). \nonumber
\end{align}

\noindent For second-order approximation, 
\begin{align}
    & \mu_m = h_1(\EE_{q}(\w_{\setminus m})) + \frac{1}{2} tr\left(\mathrm{var}_{q}(\w_{\setminus m})h_3(\EE_{q}(\w_{\setminus m}))\right), \nonumber \\
    & v_m = h_2(\EE_{q}(\w_{\setminus m})) + \frac{1}{2} tr\left(\mathrm{var}_{q}(\w_{\setminus m})h_4(\EE_{q}(\w_{\setminus m}))\right). \nonumber
\end{align}

\subsection{Probit Regression}

In this section, we give the details of fully-factorized approximation using standard EP and conditional EP for probit regression model.

\noindent Given the observed classification instances $\X =[\x_1, \ldots, \x_n]^\top$, and binary labels  $\y = [y_1, \ldots, y_n]$,  the joint probability of the Bayesian probit model is
\begin{align}
    p(\y, \w|\X) = p(\w) \prod_{i=1}^n \psi\left((2y_i - 1)\w^\top \x_i\right), \nonumber 
\end{align} 
where $p(\w) = \N(\w|\0,\lambda \I)$ and $\psi(\cdot)$ is the cumulative density function (CDF) of the standard normal distribution.

\noindent Fully-factorized appproximate posterior distribution can be described as 
\begin{align}
    q(\w) \propto p(\w) \prod_{i=1}^n \prod_m\tf_{im}(w_m) \nonumber
\end{align}
where $\tf_{im}(w_m) = \N(w_m|\mu_{im}, v_{im})$.

\noindent The cavity distribution and tilted distribution are as follows: 
\begin{align}
    & q^{\setminus i}(\w) \propto p(\w) \prod_{j \neq i} \prod_m\tf_{jm}(w_m) = \prod_m \N(w_m|\mu_m^{\setminus i}, v_m^{\setminus i}), \nonumber \\
    & \hat{p}_i(\w) \propto  \psi\left((2y_i - 1)\w^\top \x_i\right) q^{\setminus i}(\w), \nonumber
\end{align}
where
\begin{align}
    &v_m^{\setminus i} = \frac{1}{1/\lambda + \sum_{j \neq i} 1/v_{jm}}, \nonumber \\
    &\mu_m^{\setminus i} = v_m^{\setminus i}\sum_{j \neq i} \frac{\mu_{jm}}{v_{jm}}. \nonumber
\end{align}

\subsubsection{Standard EP for Probit Regression}

To apply standard EP method, it is tricky to compute the moments of unnormalized tilted distribution.

\noindent Let
\begin{align}
    Z_i = & \int \psi\left((2y_i - 1)\w^\top \x_i\right) \prod_m \N(w_m|\mu_m^{\setminus i}, v_m^{\setminus i}) \nonumber \\
        = & \psi\left(\frac{(2y_i - 1)\sum_m x_{im}\mu_m^{\setminus i}}{\sqrt{1 + x_{im}^2 v_m^{\setminus i}}}\right) \nonumber.
\end{align}

\noindent The mean and variance of the tilted distribution can be derived as 
\begin{align}
    & \mu_m = \mu_m^{\setminus i} + v_m^{\setminus i}\frac{\partial \log Z_i}{\partial \mu_m^{\setminus i}}, \nonumber \\
    & v_m = v_m^{\setminus i} - {v_m^{\setminus i}}^2\left( (\frac{\partial \log Z_i}{\partial \mu_m^{\setminus i}})^2 - 2 \frac{\partial \log Z_i}{\partial v_m^{\setminus i}} \right). \nonumber
\end{align}

\noindent By subtracting off the cavity distribution, we can get updated $v_{im}$ and $\mu_{im}$,
\begin{align}
    &v_{im}^* = \left(\frac{1}{v_m} - \frac{1}{v_m^{\setminus i}} \right)^{-1} \nonumber \\
    &\mu_{im}^* = v_{im}^* \left( \frac{\mu_m}{v_m} - \frac{\mu_m^{\setminus i}}{v_m^{\setminus i}} \right). \nonumber
\end{align}

\subsubsection{Conditional EP for Probit Regression}

The conditional tilted distribution for probabit model is
\begin{align}
    \hat{p}_i(w_m|{\w_i}_{\setminus i}) \propto \psi\left((2y_i - 1)\w^\top \x_i\right) \N(w_m|\mu_m^{\setminus i}, v_m^{\setminus i}). \nonumber
\end{align}

\noindent We use the similar procedure applied in previous section to compute the conditional moments,
\begin{align}
    & \EE(w_m|{\w_i}_{\setminus i}) = \mu_m^{\setminus i} + v_m^{\setminus i}\frac{\partial \log Z_{im}}{\partial \mu_m^{\setminus i}}, \nonumber \\
    & \mathrm{cov}(w_m|{\w_i}_{\setminus i}) = v_m^{\setminus i} - {v_m^{\setminus i}}^2\left( (\frac{\partial \log Z_{im}}{\partial \mu_m^{\setminus i}})^2 - 2 \frac{\partial \log Z_{im}}{\partial v_m^{\setminus i}} \right), \nonumber
\end{align}
where
\begin{align}
    Z_{im} =& \int \psi\left((2y_i - 1)\w^\top \x_i\right) \N(w_m|\mu_m^{\setminus i}, v_m^{\setminus i}) \nonumber \\
        =& \psi\left( \frac{(2y_i - 1)(x_{im}\mu_{m}^{\setminus i} + \w_{\setminus m}^\top {\x_i}_{\setminus m})}{\sqrt{1 + x_{im}^2 v_m^{\setminus i}}} \right). \nonumber
\end{align}

\noindent By denoting
\begin{align}
    &\N = \N\left(\frac{(2y_i - 1)(x_{im}\mu_{m}^{\setminus i} + \w_{\setminus m}^\top {\x_i}_{\setminus m})}{\sqrt{1 + x_{im}^2 v_m^{\setminus i}}}|0,1\right), \nonumber \\
    &\psi = \psi\left( \frac{(2y_i - 1)(x_{im}\mu_{m}^{\setminus i} + \w_{\setminus m}^\top {\x_i}_{\setminus m})}{\sqrt{1 + x_{im}^2 v_m^{\setminus i}}} \right), \nonumber \\
    &c_1 = \frac{(2y_i - 1)}{\sqrt{1 + x_{im}^2 v_m^{\setminus i}}}, \nonumber \\
    &c_2 = x_{im}\mu_{m}^{\setminus i} + \w_{\setminus m}^\top {\x_i}_{\setminus m}) \nonumber,
\end{align}
we have the Hessian matrix of conditional moments derived w.r.t $\w_{\setminus m}$,
\begin{align}
    &\nabla \nabla \EE(w_m|{\w_i}_{\setminus i}) = T_1 c_1^3 v_m^{\setminus i} x_{im} {\x_i}_{\setminus m}{\x_i}_{\setminus m}^\top, \nonumber \\
    &\nabla \nabla \mathrm{cov}(w_m|{\w_i}_{\setminus i}) = T_2 c_1^4 {v_m^{\setminus i}}^2 x_{im}^2{\x_i}_{\setminus m}{\x_i}_{\setminus m}^\top, \nonumber
\end{align}
where
\begin{align}
    &T_1 = (c_1^2c_2^2 - 1)\frac{\N}{\psi} + 3c_1c_2\frac{\N^2}{\psi^2} + 2\frac{\N^3}{\psi^3}, \nonumber \\
    &T_2 = c_1c_2(3-c_1^2c_2^2)\frac{\N}{\psi} + (4 - 7c_1^2c_2^2)\frac{\N^2}{\psi^2} \nonumber \\
    &\quad - 12c_1c_2\frac{\N^3}{\psi^3} - 6 \frac{\N^4}{\psi^4}. \nonumber 
\end{align}

\noindent After computing the mean $\mu_m$ and variance $v_m$ of tilted distribution, $v_{im}$ and $\mu_{im}$ can be updated as 
\begin{align}
    &v_{im}^* = \left(\frac{1}{v_m} - \frac{1}{v_m^{\setminus i}} \right)^{-1} \nonumber \\
    &\mu_{im}^* = v_{im}^* \left( \frac{\mu_m}{v_m} - \frac{\mu_m^{\setminus i}}{v_m^{\setminus i}} \right). \nonumber
\end{align}

\noindent Denote
\begin{align}
    & h_1(\w_{\setminus m}) = \EE(w_m|\w_{\setminus m}), \nonumber \\
    & h_2(\w_{\setminus m}) = \mathrm{var}(w_m|\w_{\setminus m}), \nonumber \\
    & h_3(\w_{\setminus m}) = \nabla \nabla \EE(w_m|\w_{\setminus m}), \nonumber \\
    & h_4(\w_{\setminus m}) = \nabla \nabla \mathrm{var}(w_m|\w_{\setminus m}). \nonumber
\end{align}

\noindent For fisrt-order approximation, 
\begin{align}
    & \mu_m = h_1(\EE_{q}(\w_{\setminus m})), \nonumber \\
    & v_m = h_2(\EE_{q}(\w_{\setminus m})), \nonumber
\end{align}

\noindent For second-order approximation, 
\begin{align}
    & \mu_m = h_1(\EE_{q}(\w_{\setminus m})) + \frac{1}{2} tr\left(\mathrm{var}_{q}(\w_{\setminus m})h_3(\EE_{q}(\w_{\setminus m}))\right), \nonumber \\
    & v_m = h_2(\EE_{q}(\w_{\setminus m})) + \frac{1}{2} tr\left(\mathrm{var}_{q}(\w_{\setminus m})h_4(\EE_{q}(\w_{\setminus m}))\right), \nonumber
\end{align}
where $\EE_{q}(\cdot)$ and $\mathrm{var}_{q}(\cdot)$ are the expectation and covariance w.r.t $q(\w_{\setminus m})$.

\section{Experiment}
\subsection{Baysian Probit and Logistic Regression}

\begin{table*}
	\begin{subtable}{\textwidth}
    \centering
	\small
	\begin{tabular}[c]{cccccc}
		\hline\hline
		Dataset & CEP-1 & CEP-2 & LP &  EP & VB \\
		\hline 
		australian & $\mathbf{0.883 \pm 	0.009}$ & $\mathbf{0.883 \pm 0.009}$ & $\mathbf{0.883 \pm 	0.009 }$ & $0.880 \pm	0.009$ & $0.878 \pm 0.010$ \\
		breast & $0.686 \pm	0.016$ & $ \mathbf{0.690 \pm  0.016}$ & $0.681 \pm  0.020$ & $0.677 \pm	0.020$ & $0.672\pm	0.022$ \\
		crab & $\mathbf{0.995 \pm	0.002}$ & $\mathbf{0.995 \pm	0.002}$ &
		$\mathbf{0.995 \pm	0.002}$ &
		$\mathbf{0.995 \pm	0.002}$ &		
		 $\mathbf{0.995 \pm	0.002}$ \\
		 ionos & $0.926\pm	0.006$ & $\mathbf{0.929 \pm	0.006}$ & 
		 $0.896 \pm	0.006$ & $0.903 \pm 0.005$ & $0.883 \pm 	0.010$ \\
		 pima & $\mathbf{0.825 \pm	0.004}$ & 
		 $\mathbf{0.825 \pm	0.004}$ & 
		 $\mathbf{0.825 \pm	0.004}$ &
		 $\mathbf{0.825\pm	0.004}$ &
		 $\mathbf{0.825 \pm	0.004}$ \\
		 sonar & $0.813 \pm	0.023$ & $\mathbf{0.831 \pm	0.022}$ &  $0.804 \pm	0.020$  &  $0.813 \pm	0.022$ & $0.797\pm	0.021$\\
		\hline 
	\end{tabular}
	\caption{Bayesian probit regression} 
    \end{subtable}
\begin{subtable}{\textwidth}
	\centering
	\small
	\begin{tabular}[c]{cccccc}
		\hline\hline
		Dataset & CEP-1 & CEP-2 & LP &  EP & VB \\
		\hline 
		australian & $\mathbf{0.873 \pm 	0.008}$ & $0.873\pm 0.010$  & $0.873 \pm 	0.009$ & $0.873 \pm	0.010$ & $0.873 \pm 0.009$ \\
		breast & $0.675 \pm	0.021$ & $\mathbf{0.683 \pm 	0.020}$  & $0.675 \pm 	0.021$ & $0.677 \pm	0.020$ & $0.676\pm	0.019$ \\
		crab & $\mathbf{0.993 \pm	0.001}$ & $0.992 \pm	0.001$ &
		 $\mathbf{0.993 \pm	0.001}$ & $\mathbf{0.993 \pm	0.001}$ &
		$\mathbf{0.993 \pm	0.001}$ \\
		ionos & $0.912 \pm	0.011$ & $\mathbf{0.925 \pm 0.010}$ &   $0.912 \pm	0.012$ & $0.914 \pm 0.010$ &
		$0.908 \pm	0.010$ \\
		pima & $0.830 \pm	0.005$ & 
		$\mathbf{0.831 \pm	0.005}$ &
		$\mathbf{0.831 \pm	0.005}$ &
		$0.830\pm  0.005$ & 
		$0.830 \pm	0.005$ \\
		sonar & $0.820 \pm	0.015$ &
		$\mathbf{0.831	\pm 0.016}$ &
		$0.820 \pm	0.016$ & $0.827 \pm	0.016$ &
		$0.825\pm	0.014$\\ 
		\hline 
    \end{tabular}
	\caption{Bayesian logistic regression} 
\end{subtable}
\centering
\caption{Average test AUC on six real datasets.}
\label{tb:auc}
\end{table*}

Table \ref{tb:auc} lists the results of test AUC on six real-world datasets from UCI machine learning repository\footnote{\url{https://archive.ics.uci.edu/ml/index.php}}, \textit{australian}, \textit{breast}, \textit{crab},\textit{ionos},  \textit{pima} and \textit{sonar}. As we can see,
in Baysian probit regression model, from the perspective of AUC, CEP-1 and CEP-2 can always get a better or same performance than standard EP and CEP-2 can always get the best performance on all six datasets. In Bayesian logistic regression model, CEP-1 and CEP-2 have similar performance with standard EP. CEP-2 get the highest AUC on
\textit{brerast}, \textit{ionos}, \textit{pima} and \textit{sonar} and CEP-1 get the highest AUC on the other two datasets, \textit{australian} and \textit{crab}.

\subsection{Tensor Decomposition}

\begin{figure*}
	\centering
	\setlength\tabcolsep{0.1pt}
	\begin{tabular}[c]{cc}
		\begin{subfigure}{0.25\textwidth}
			\centering
			\includegraphics[width=\textwidth]{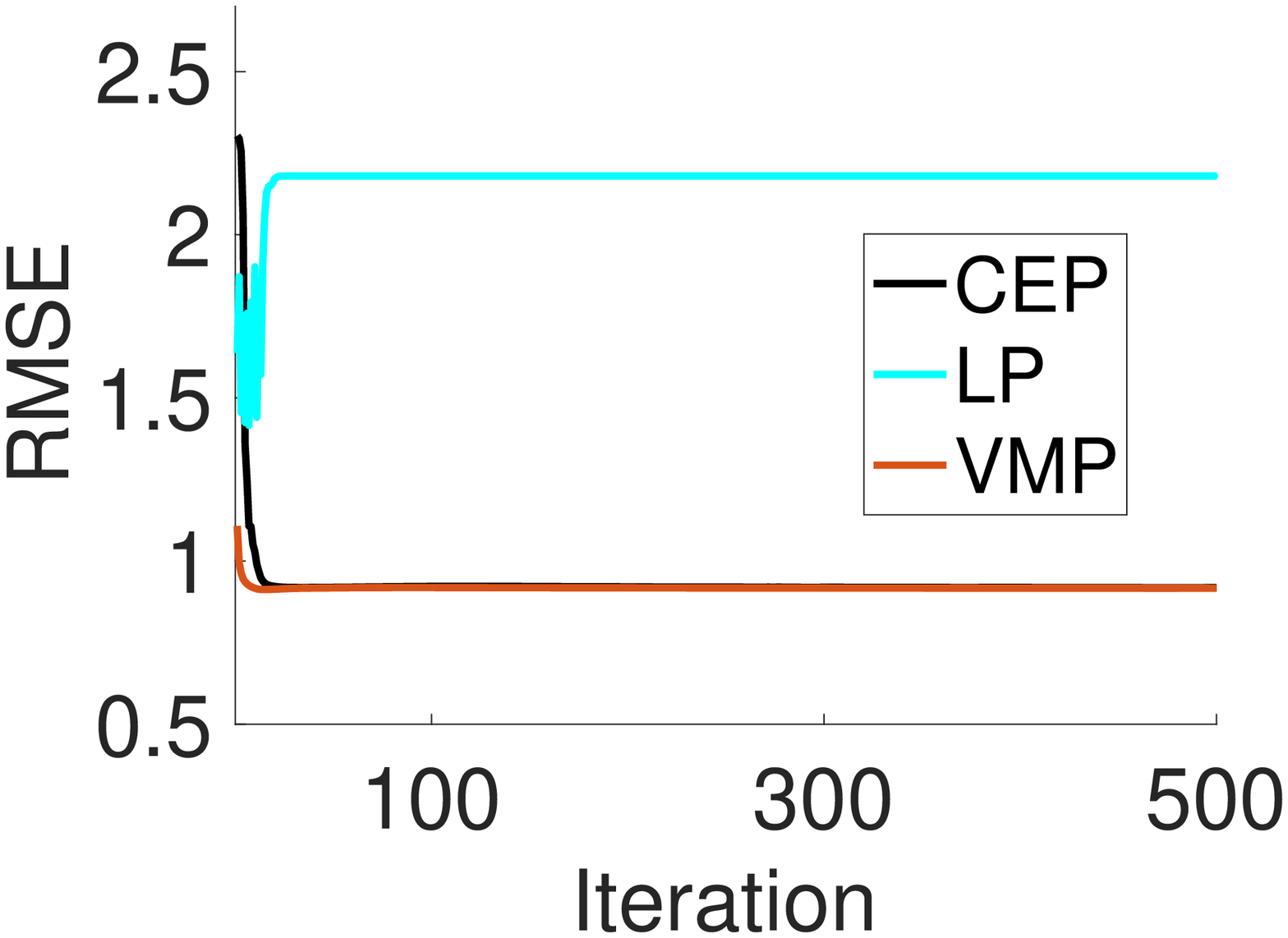}
			\caption{\textit{Alog}}
        \end{subfigure}
		&
		\begin{subfigure}{0.25\textwidth}
			\centering
			\includegraphics[width=\textwidth]{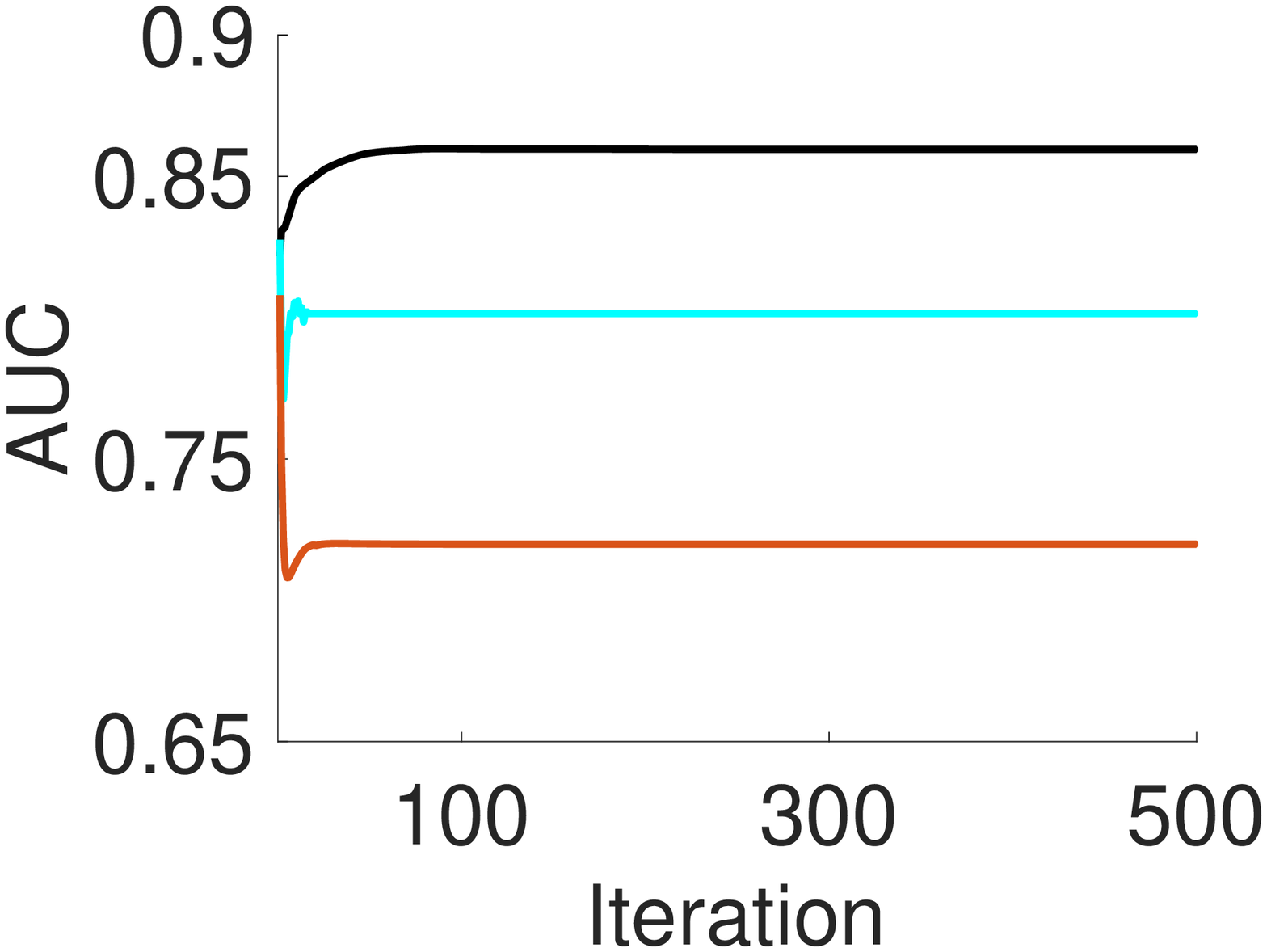}
			\caption{\textit{Enron}}
		\end{subfigure}  
	\end{tabular}
	\caption{Average prediction accuracy \textit{v.s.} running iteration. The number of latent factors is $3$.} 
	\label{fig:tf-iter-r3}
	\vspace{-0.2in}
\end{figure*}

\begin{figure*}
	\centering
	\setlength\tabcolsep{0.1pt}
	\begin{tabular}[c]{cc}
		\begin{subfigure}{0.25\textwidth}
			\centering
			\includegraphics[width=\textwidth]{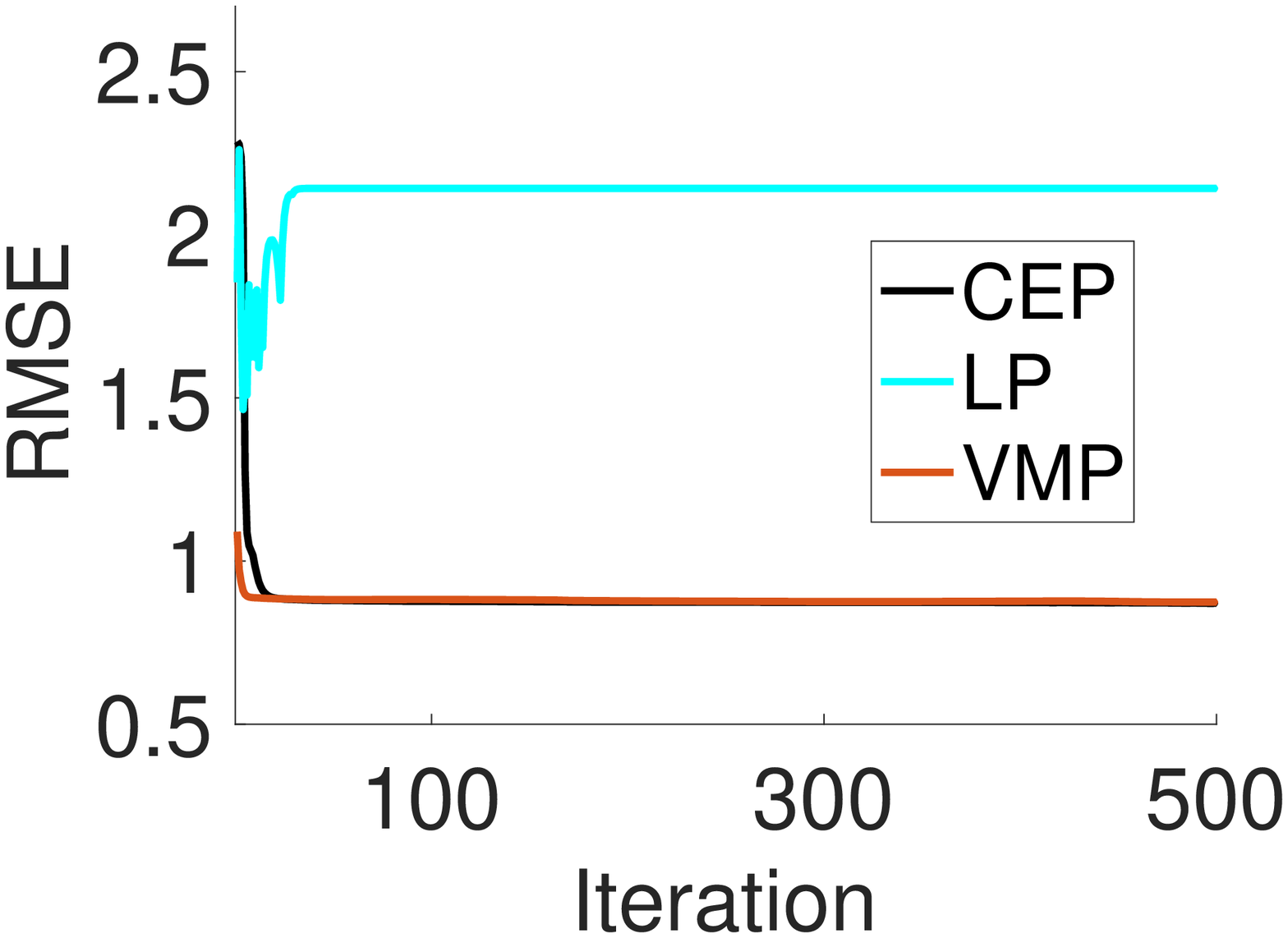}
		\end{subfigure}
		&
		\begin{subfigure}{0.25\textwidth}
			\centering
			\includegraphics[width=\textwidth]{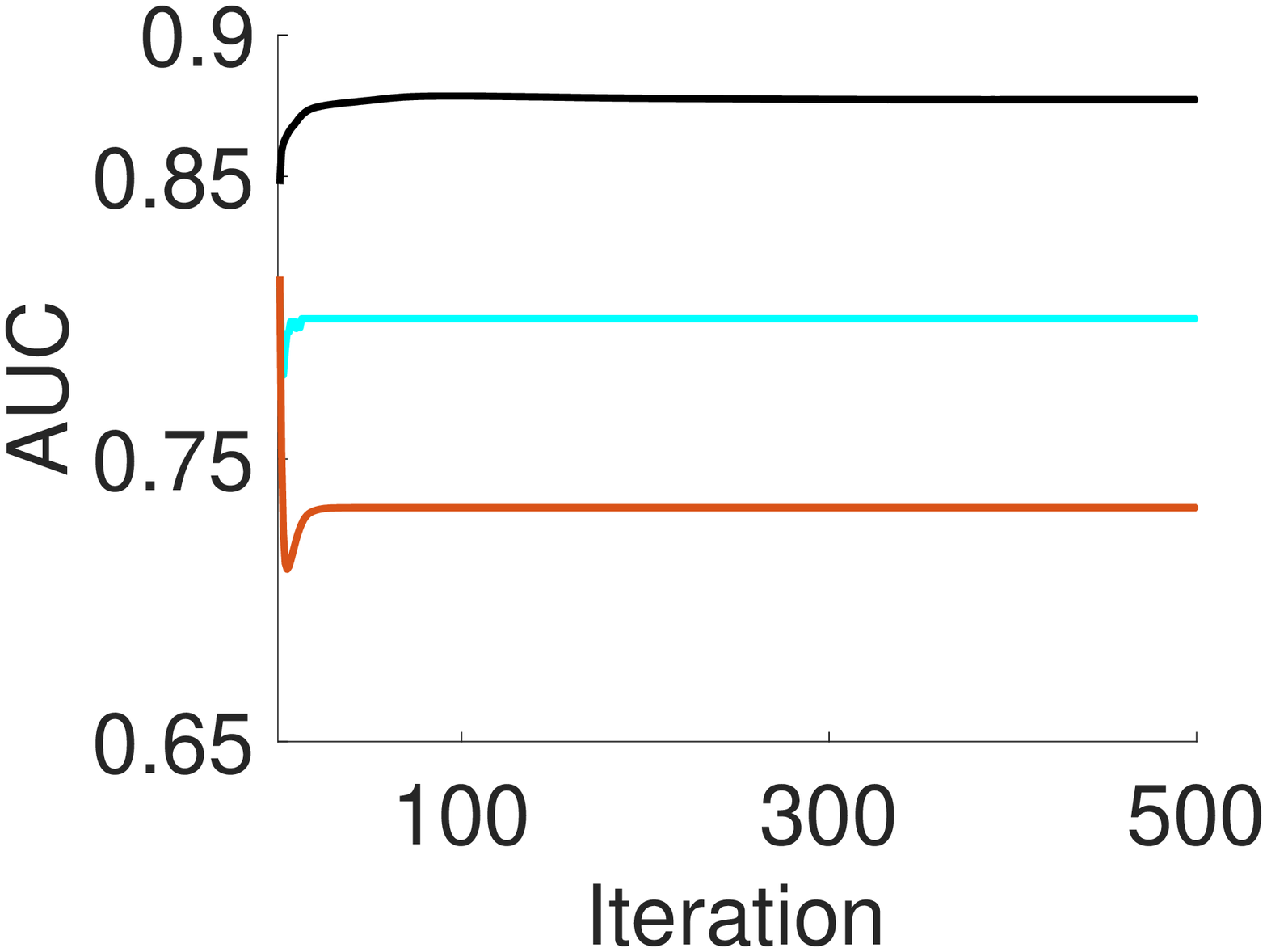}
		\end{subfigure}  
	\end{tabular}
	\caption{Average prediction accuracy \textit{v.s.} running iteration. The number of latent factors is $5$.} 
	\label{fig:tf-iter-r5}
	\vspace{-0.2in}
\end{figure*}

\begin{figure*}
	\centering
	\setlength\tabcolsep{0.1pt}
	\begin{tabular}[c]{cc}
		\begin{subfigure}{0.25\textwidth}
			\centering
			\includegraphics[width=\textwidth]{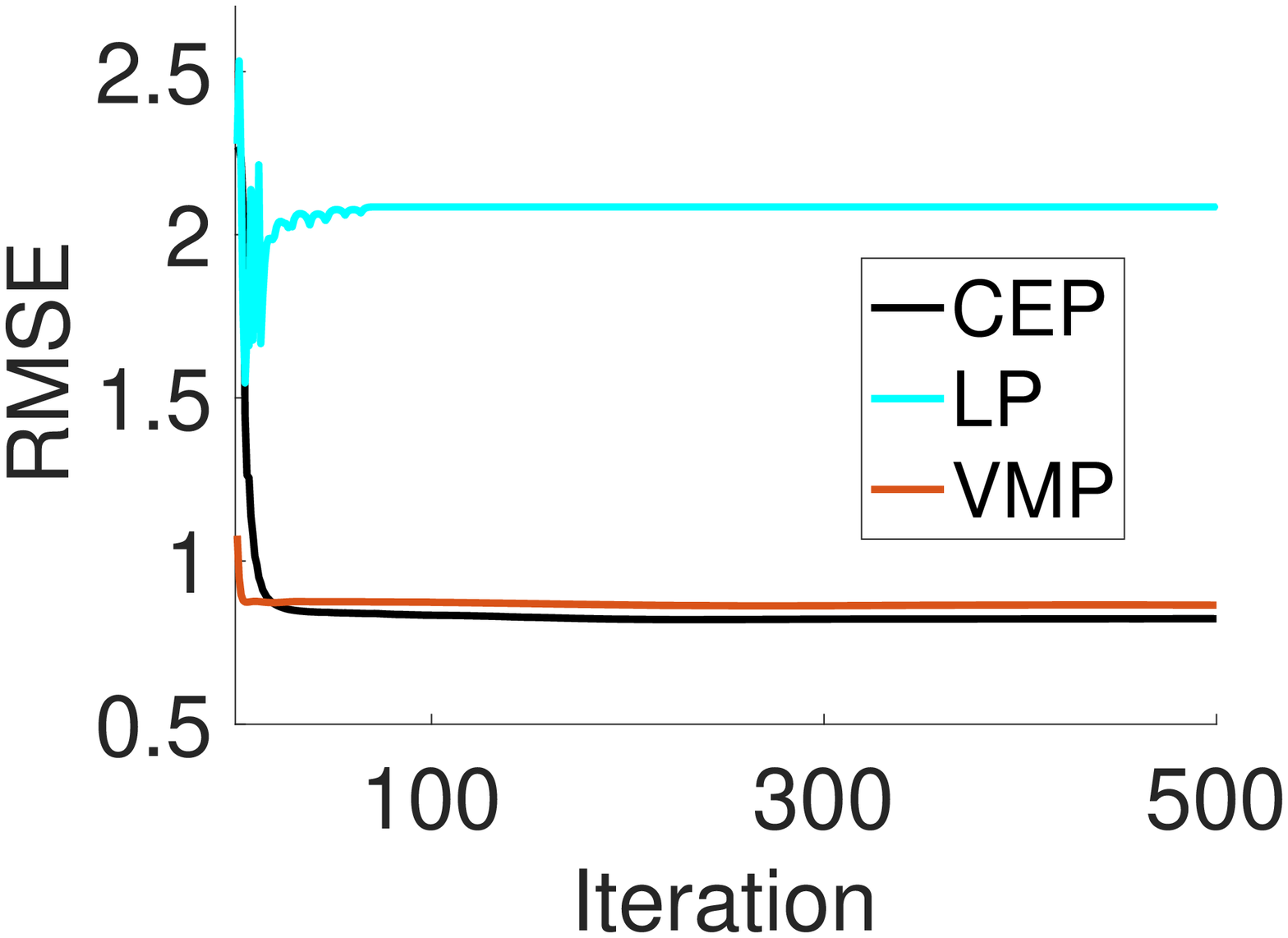}
			\caption{\textit{Alog}}
		\end{subfigure}
		&
		\begin{subfigure}{0.25\textwidth}
			\centering
			\includegraphics[width=\textwidth]{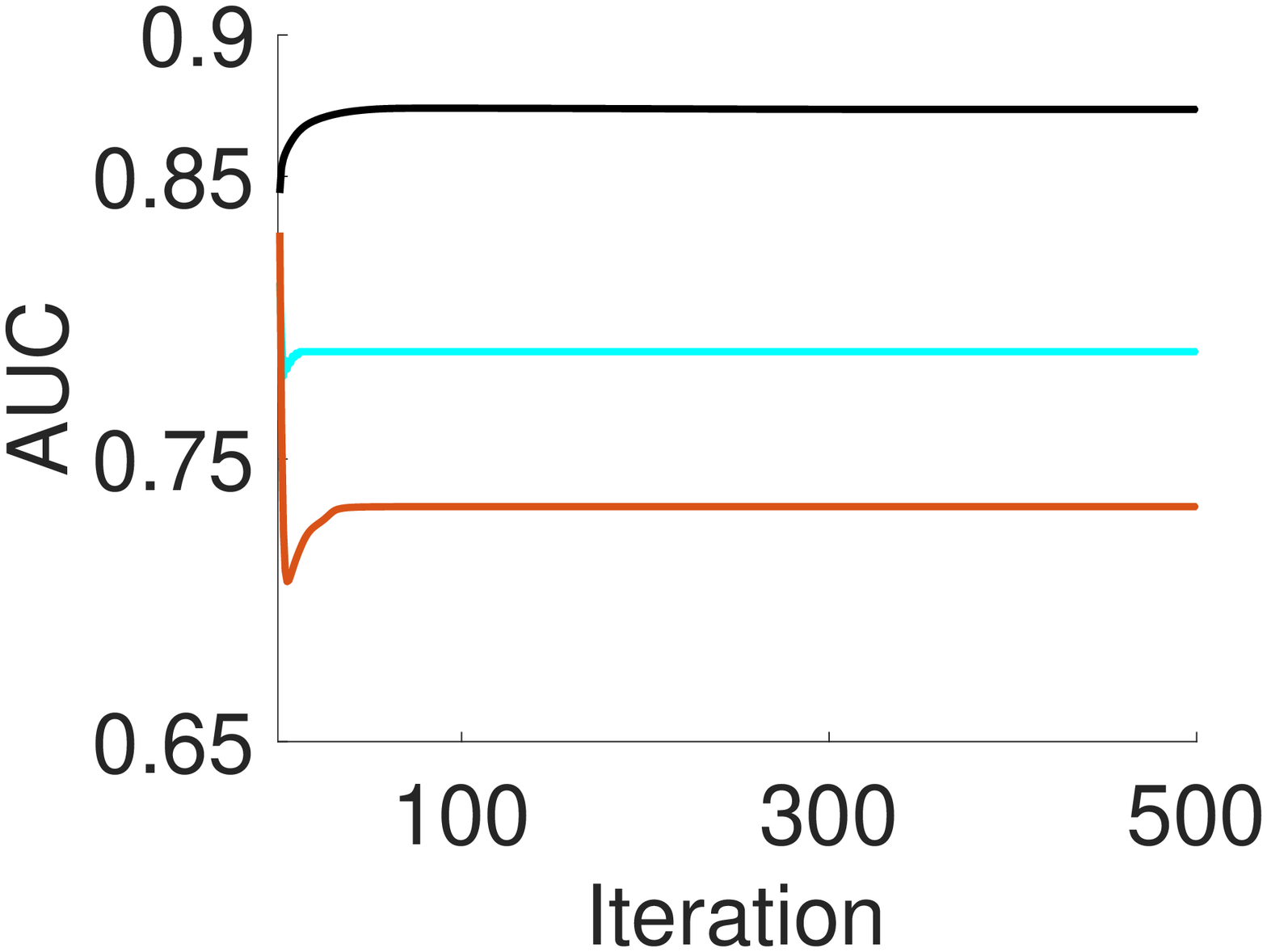}
			\caption{\textit{Enron}}
		\end{subfigure}  
	\end{tabular}
	\caption{Average prediction accuracy \textit{v.s.} running iteration. The number of latent factors is $8$.} 
	\label{fig:tf-iter-r8}
	\vspace{-0.2in}
\end{figure*}
We show in Fig. \ref{fig:tf-iter-r3}, Fig. \ref{fig:tf-iter-r5} and Fig. \ref{fig:tf-iter-r8} how the predictive performance of CEP, LP and VMP varied along with the running iterations when the number of latent factors is 3, 5 and 8 respectively. As we can see, the prediction accuracy of all the three methods converges quickly and keeps stable with more iterations.
Results on \textit{Alog} show that CEP can achieve a similar or even better performance than VMP, while LP has the worst performance. However, results on \textit{Eron} show that CEP achieves the highest AUC and LP performs better than VMP. In both cases, CEP can get a better performance than the other two methods.

\begin{table*}
	\begin{tabular}[c]{cccccc}
		\hline\hline
		Dataset & Feature Size & Sample Size & Link \\
		\hline 
		australian & 14 & 690 & https://archive.ics.uci.edu/ml/datasets/Statlog+\%28Australian+Credit+Approval\%29 \\
		breast & 9 & 286 & https://archive.ics.uci.edu/ml/datasets/Breast+Cancer \\
		crab & 7 & 200 & https://vincentarelbundock.github.io/Rdatasets/csv/MASS/crabs.csv \\
		ionos & 34 & 351 & https://archive.ics.uci.edu/ml/datasets/Ionosphere \\
		pima & 7 & 332 & https://vincentarelbundock.github.io/Rdatasets/csv/MASS/Pima.te.csv\\
		sona & 60 & 208 & https://archive.ics.uci.edu/ml/datasets/Connectionist+Bench+\%28Sonar\%2C+Mines+vs.+Rocks\%29 \\
		\hline 
	\end{tabular}
	\caption{Dataset Description} 
    \label{tb:dataset}
\end{table*}

\end{document}